\pdfoutput=1

\documentclass[11pt]{article}

\usepackage[]{ACL2023}

\usepackage{times}
\usepackage{latexsym}

\usepackage[T1]{fontenc}

\usepackage[utf8]{inputenc}

\usepackage{microtype}

\usepackage{inconsolata}
\usepackage[textsize=scriptsize]{todonotes}

\usepackage{graphicx}
\usepackage{makecell}
\newcommand{\eat}[1]{}
\usepackage{subcaption}
\usepackage{amsmath}
\usepackage{tablefootnote}
\usepackage{threeparttable}

\newcommand{\edit}[1]{{#1}}

\newcommand{\dataset}[0]{ParRoT}
\newcommand{\method}[0]{ParRoT-Con}
\newcommand{\mm}[0]{parts mental model}
\newcommand{\mms}[0]{parts mental models}

\newcommand{\Mms}[0]{Parts mental models}

\newenvironment{enu}{                   
     \parskip 0cm \begin{list}{}{\parsep 0cm \itemsep 0cm \topsep 0cm}}{
       \end{list}} 
       
\title{Do language models have coherent mental models of everyday things?}

\author{
Yuling Gu \and Bhavana Dalvi Mishra \and Peter Clark \\
Allen Institute for AI, Seattle, WA \\
\texttt{\{yulingg,bhavanad,peterc\}@allenai.org} 
}

\begin{document}
\maketitle
\begin{abstract}
When people think of everyday things like an egg, they typically have a mental image associated with it. This allows them to correctly judge, for example, that ``the \edit{yolk} surrounds the shell'' is a false statement. Do language models similarly have a coherent picture of such everyday things? To investigate this, we propose a benchmark dataset consisting of 100 everyday things, their parts, and the relationships between these parts, expressed as 11,720 ``X relation Y?'' true/false questions. Using these questions as probes, we observe that state-of-the-art pre-trained language models (LMs) like GPT-3 and Macaw have fragments of knowledge about these everyday things, but do not have fully coherent ``parts mental models'' (54-59\% accurate, 19-43\% conditional constraint violation). We propose an extension where we add a constraint satisfaction layer on top of the LM's raw predictions to apply commonsense constraints. As well as removing inconsistencies, we find that this also significantly improves accuracy (by 16-20\%), suggesting how the incoherence of the LM's pictures of everyday things can be significantly reduced.\footnote{\edit{We make our data and code publicly available at \url{https://github.com/allenai/everyday-things}.}}

\end{abstract}

\begin{figure}[th!]
\centering
      \includegraphics[width=\columnwidth]{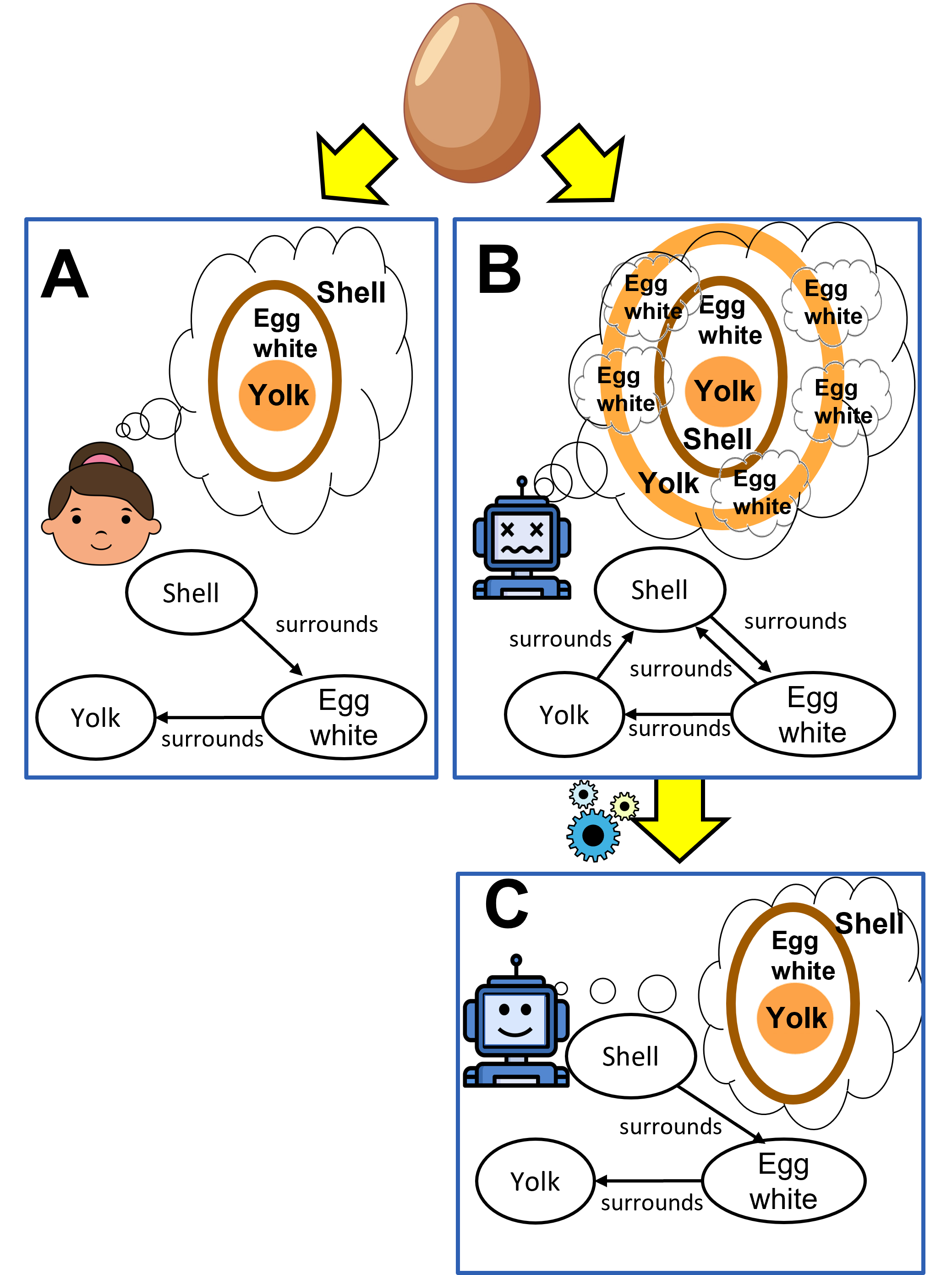}
\caption{While humans appear to have coherent mental pictures of everyday things (e.g., an egg, {\bf A}),
our question-asking probes suggest that LMs do not (e.g., one LM answered that the egg white both
surrounds and is surrounded by the shell, {\bf B}). This model incoherence can be
reduced by applying commonsense constraints (e.g., surrounds is asymmetric), resulting in
a more coherent parts model ({\bf C}).}
\label{fig:task-pic}
\end{figure}

\section{Introduction}
Psychologists and cognitive scientists hypothesize that humans develop mental models of the world, namely internal, conceptual representations of the environment which we base our decisions and actions on \cite{Ha2018WorldM, Jonassen1996MentalMK}. \citet{Hespos-nature-2004} observed that 5-month-old human infants exhibit understanding of mechanical properties of objects in terms of arrangements and motions of surfaces, well before they can understand language.  Drawing loosely on this idea, but without making any claims about how LMs
reason internally \cite{lm-humans-1,lm-humans-2}, we  investigate if pre-trained language models 
show evidence of coherent internal representations of everyday things, analogous to human mental models, via probing.
We focus on mental models in the context of ordinary objects that we encounter in our everyday lives. 
Such commonsense knowledge helps us understand how these everyday things work and how to interact with them. For example, when someone tries to make a fried egg, they know that it has a shell and that it can be cracked open to reveal the egg white and yolk inside. However, if a system does not have a coherent picture of such everyday things, thinking that the egg yolk surrounds the shell, then it might have to resort to ridiculous approaches such as trying to scrape the egg yolk off the shell into the pan. 

\begin{figure*}
\captionsetup[subfigure]{labelformat=empty}
\centering
\begin{subfigure}{.55\textwidth}
  \centering
  \includegraphics[width=\linewidth]{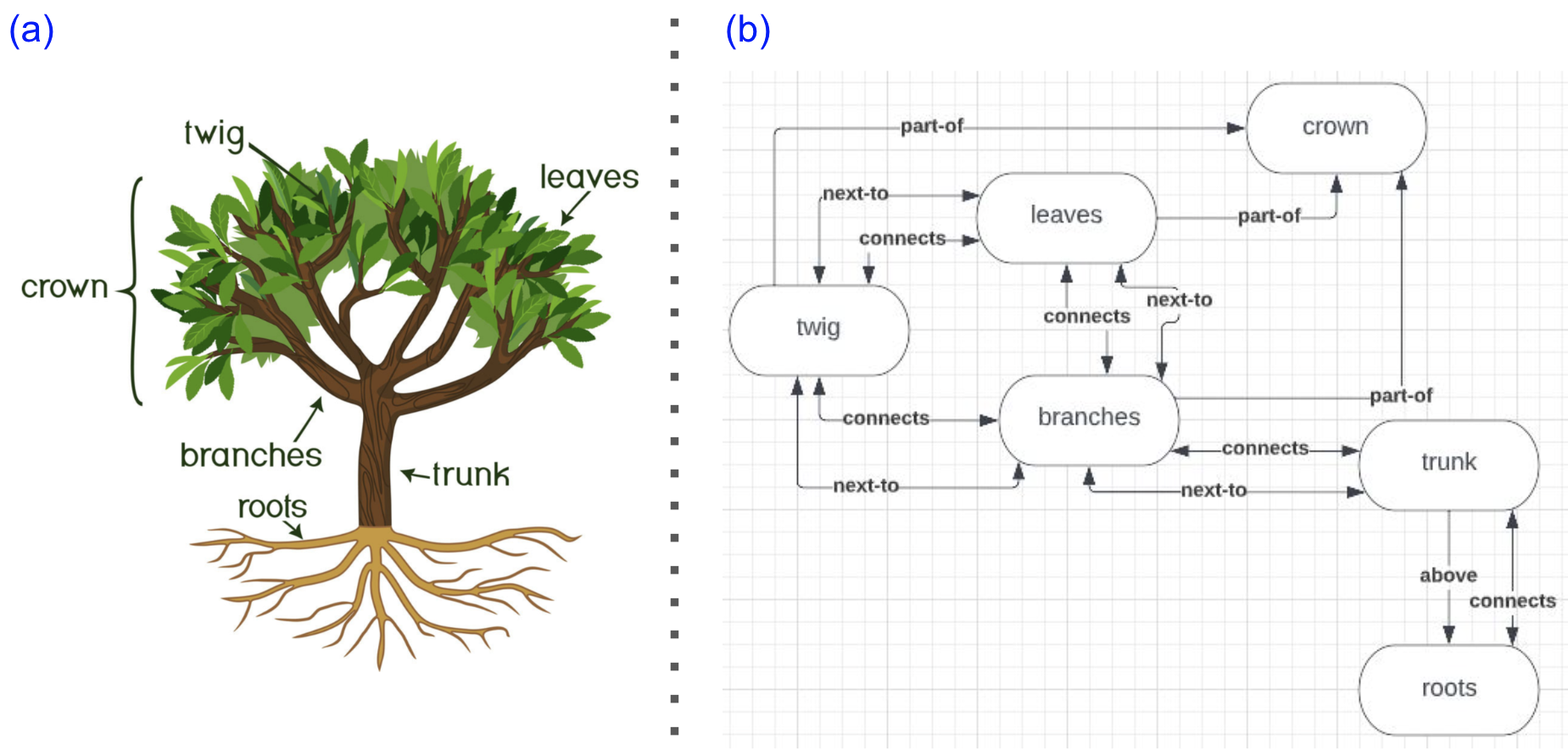}
  \caption{Tree}
  \label{fig:sub1}
\end{subfigure}%
\begin{subfigure}{0.46\textwidth}
  \centering
  \includegraphics[width=\linewidth]{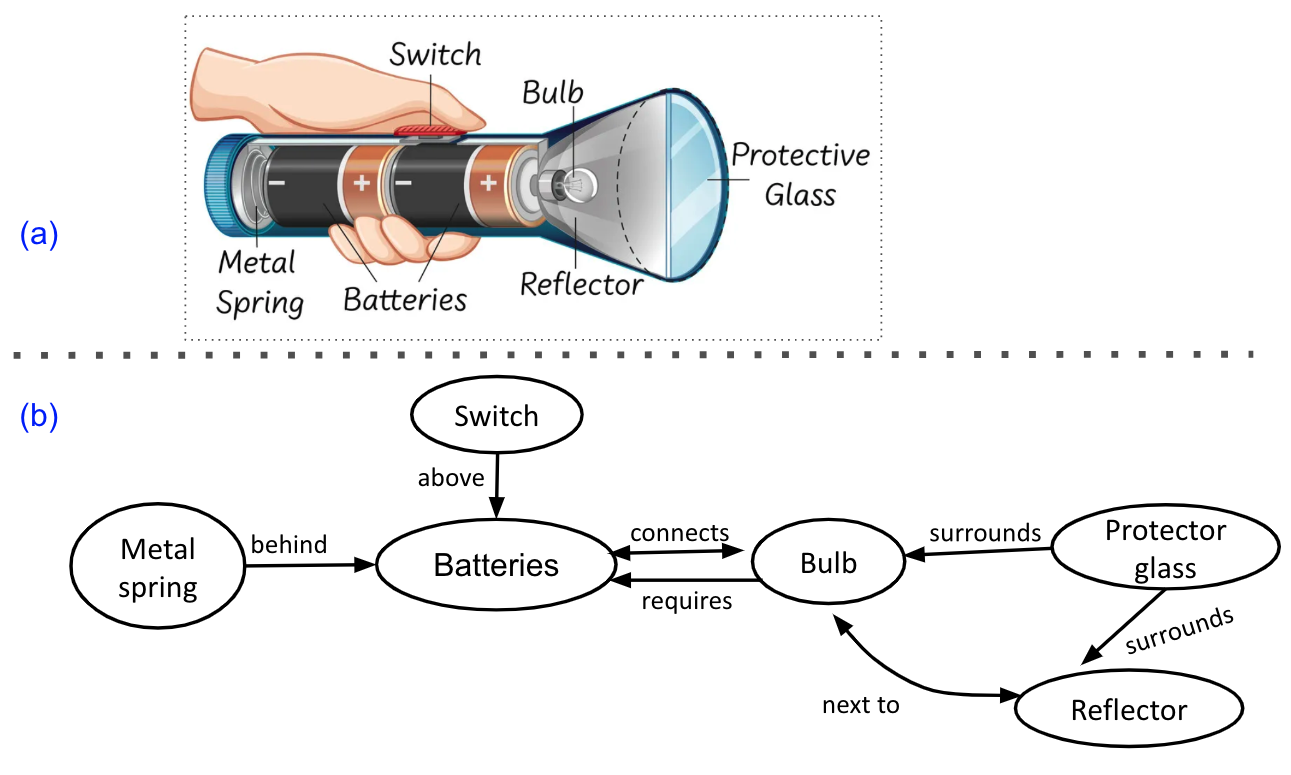}
  \caption{Flashlight}
  \label{fig:sub2}
\end{subfigure}
\caption{Our everyday things dataset, \dataset, covers different entities, both natural (e.g. tree) and man-made (e.g. flashlight). Above are two examples of such everyday things. In each case, we show a (a) diagram of the entity, and (b) parts graph of the everyday thing drawn by crowdworkers. The parts graphs illustrate how our dataset contains a variety of relations between parts.}
\label{fig:example_mm}
\end{figure*}

We explore a first version of this, in which we consider only knowledge about
an object's parts and their relationships. We refer to this knowledge as a \mm.
We first create a benchmark dataset of 100 everyday things, by asking human annotators to draw a graph representing their parts mental model (e.g., Figure \ref{fig:example_mm}) depicting the parts of an everyday thing, spatial relationships, connections between its parts and functional dependencies (if any).
Then we probe two representative state-of-the-art LMs with questions about these everyday things. We find that the LMs' \mms{}  are
generally of poor quality.  Further, model predictions can violate basic consistency constraints e.g. transitivity. To alleviate this, we apply constraint reasoning to derive more accurate and consistent mental models of everyday things, correcting some of the LMs' original inconsistencies. This is illustrated in Figure \ref{fig:task-pic}.

Our contributions are:
\begin{enu}
\item[1.] We present a benchmark dataset of \mms{} consisting of 100 everyday things, 2.2K parts and 11.7K relationships.
\item[2.] We show that SOTA LMs like GPT-3 and Macaw are poor at answering relationship queries between parts of everyday things. The \mms{} derived using their predictions are only 54-59\% accurate, and significantly inconsistent (19-43\% conditional violation $\tau$).
\item[3.] We propose a neuro-symbolic method that applies constraint reasoning on top of raw LM predictions as a way of obtaining more consistent (0\% conditional violation $\tau$) and more accurate mental models (16-20\% improvement). This suggests a broader cognitive architecture
(LM + reasoner) for future systems, to better construct mental models than the LM alone.

\end{enu}

\section{Related work}
\textbf{Mental models:}
The idea of mental models \citep{JohnsonLaird1983MentalM} is not new. Many years ago, \citet{Craik1943TheNO} proposed that thinking itself is the manipulation of internal representations of the world. \citet{Craik1943TheNO} described mental models as a ‘small-scale model’ of external reality and of its own possible actions within someone’s head. Such a mental model is useful in many ways, including allowing one to try out various alternatives, make conclusions, react to future situations, learn from past events, and in general, 
improve competency. Years later, when \citet{JohnsonLaird2006how-we-reason} outlined the mental processes that underlie human reasoning, he based his discussion on the fundamental assumption that human beings can construct internal representations of spatial layouts, and specified mental models to be iconic. In his words, a mental model's ``parts and the relations among them correspond to the parts of the layout and the relations among them.''
While coherent internal representations of spatial layouts are crucial for human reasoning, their role, coherence, and even existence in LMs have not been systematically explored. 
In this work, we try to bridge this gap by proposing a benchmark dataset and methodology to compare human internal representations of spatial layouts of everyday things with those of LMs.

\noindent \textbf{Prior datasets:} Prior works on reasoning about object/body parts include \citet{li2019hake} which focused on human body parts and human interaction with other objects. 
The PTR benchmark \citep{Hong-NEURIPS2021-ptr} is a QA dataset about objects and their parts, combining 5 everyday things: chair, table, bed, refrigerator, and cart, to create questions across 70K different scenes.
\citet{ji2022abstract} used tangram puzzles to analyze shape naming, part naming and segmentation divergence across participants when they see a certain shape.
Contributing to this existing body of datasets, the dataset we introduce serves as a resource for researchers to study canonical \mms{} for a wide variety of everyday things, focusing on relationships between parts of objects, which is fundamental to how humans think and interact with these things.

\noindent \textbf{Large language models:}
Despite recent advances in LMs, studies suggest that they still struggle at reasoning with real-world entities and concepts. \citet{Bisk_Zellers_Lebras_Gao_Choi_2020} found that when LMs answer questions involving physical commonsense reasoning, their performance at that time was near chance level for questions involving spatial relations like ``top'' and ``bottom.'' \citet{sahu2022unpacking} demonstrated the lack of conceptual consistency in LMs by correlating models' answers on commonsense reasoning questions (CSQA dataset) and their answers on associated conceptual questions from ConceptNet knowledge base. 
To improve existing systems, progress has been made such as by imposing constraints with neuro-symbolic approaches \citep{dualsys2021nye, mitchell2022enhancing} and incorporating both textual and visual information \citep{dan-etal-2020-understanding}. Inspired by recent progress, we propose a constraint reasoning method that applies hard commonsense constraints (e.g., if `A above B' is \textit{True} then `A below B' cannot be \textit{True}) on top of raw LM predictions to produce more accurate and consistent mental models of everyday things.

\section{\Mms{} and Task}

We define ``\mm{}'' for everyday things in this section. Then in the rest of the paper, we describe how we collect a dataset for them, measure LMs' coherence on them, and finally apply external reasoning to improve the accuracy and consistency of LMs' \mm{}.

Here, we use \mm{} to mean a parts-focused subset of a complete mental model of an entity.  We represent a \mm{} as a directed graph where parts of the everyday thing form the nodes of this graph and these nodes are connected with edges indicating how these parts are related to each other. Based on prior works such as \citet{rcc} and \citet{haloproject}, we selected 11 spatial orientation relations to focus on. In addition, we augmented these with relations describing connectivity and functional dependency. In total, we consider 14 relationships (across these 3 categories) between parts, listed in Table \ref{fig:4dims}.

Note that the notion of a single ``\mm{}'' for an everyday thing is 
somewhat unconstrained (e.g., which parts to pick? what version of the entity are we talking about?). To make this task more well-defined, we also provide a predefined list of parts as a guide (details in Section \ref{sec:everyday-entities}), and the task for annotators or a model is to specify relationships between them as they see appropriate,
using our ontology of relationships. 
This is important so that we can do meaningful comparisons between language models and humans' notion of \mms{} of everyday things.

\begin{table*} [h]
\centering
{\small
\setlength{\tabcolsep}{3pt}	
\begin{tabular}{l|c|cc|cc} 

                         & \makecell{Given as seed\\(unique)}  & \makecell{Annotated \\mental models} & \makecell{Avg. annotated \\ per mental model}& \makecell{Annotated + enriched (*) \\(Total)} & \edit{\makecell{Total avg. per\\ mental model \\(Total /\\\# mental models)}} \\
\hline
\# everyday things        & 100     & 100    & -   & 100    & -  \\
\# mental models          & -       & 300    & -   & 300    & -  \\
\# parts                  & 716     & 2191   & 7.30  & 2191   & 7.30 \\
\# relations (p1, rln, p2)  &   8   & 2752   & 9.17  & 11720  & 39.07 \\
\# spatial relations      &   6     &  1858  & 6.19   & 9956   & 33.19 \\
\# connectivity relation(s) &   1   & 818    & 2.73  & 1612   & 5.37 \\
\# functional relation(s)   &   1   &  76    & 0.25  & 152    & 0.51\\
\hline
\end{tabular}
}
\caption{Statistics of \dataset, our Everyday Things Dataset. *Enriched refers to implied relations, see Section \ref{sec:dataset_stats}
\label{table:dataset_stats} }
\end{table*}

\begin{table} [h]
\centering
{
\setlength{\tabcolsep}{2pt}	
\begin{tabular}{c|l} 
Type & Relations \\ \hline
\makecell{Spatial \\ orientation} & \makecell[l]{part of, has part, inside, contains, \\ in front of, behind, above, below, \\surrounds, surrounded by, next to$^*$ }\\ \hline
Connectivity & directly connected to$^*$ \\ \hline
\makecell{Functional \\dependency} & requires\tablefootnote{$A$ requires $B$ denotes $A$ cannot perform its primary function without $B$.}, required by\\
\hline
\end{tabular}
}
\caption{Relationships encoded in ``\mms{}'' of everyday things. Among these relations, `next to' and `directly connected to' relations are bi-directional, whereas the other 12 relations are uni-directional. \label{fig:4dims} }
\end{table}





Figure \ref{fig:example_mm} shows two examples of \mms{} in our dataset,
where edges encode relationships between parts. E.g., in a tree, ``trunk is above the roots''; in a flashlight, 
``bulb requires the batteries,'' etc.
Inspired by previous literature, we envision that such \mms{} 
would play a key role when one carries out daily activities involving these everyday things. 

\subsubsection*{Task}
Here we define our task: ``Construct a \mm{} for everyday things'' with the following input/output specifications:

\begin{itemize}
    \item Input: Everyday thing, Parts list, Relation vocabulary (14 relations).
    \item Output: List of tuples ($x$, $r$, $y$) where relation $r$ holds between parts $x$ and $y$.
\end{itemize}





In Section \ref{sec:dataset} we describe how we acquire a benchmark dataset by asking human annotators to carry out this task. 
Once we have collected gold-standard \mms{} for everyday things based on the human annotations, we prompt LMs for their \mms{} and evaluate how well they do on this task. Our proposed method to measure this is described in Section \ref{sec:proposed-method}. In particular, we are interested in (1) how accurate are LM-generated \mms{} when compared to gold-standard models in our dataset and (2) 
ignoring accuracy, how consistent are these generated \mms{} with respect to basic commonsense constraints? I.e., Do they at least conform to the 4 types of commonsense constraints laid out in Section \ref{sec:constrained_reasoning} e.g., `$above$' and `$below$' are inverse relations, so if the LM predicts that in a tree, (trunk is $above$ the roots) then it should also predict (roots are $below$ the trunk). 

\section{Everyday Things Dataset: ParRoT (\underline{Par}ts and \underline{R}elations  \underline{o}f \underline{T}hings) } \label{sec:dataset}
We created a dataset
of common entities that one would encounter in their daily life. For each everyday thing, our dataset (\dataset{}) contains a ``\mm{}'' in the form of a graph, which depicts parts of the entity and relational information about the parts. \edit{Such a graph encodes a parts-focused mental model of that everyday thing, potentially useful for reasoning about how the entity works and how to interact with it.}

\subsection{Everyday entities}
\label{sec:everyday-entities} 
We first compiled a list of entities from children's books, vocabulary lists (Grades 1-8), and online web search.\footnote{Appendix \ref{appendix-source-of-etc} provides more details on the source of the list of everyday things.} For the unique entities in this list, the authors 
manually filtered out those entities that are not common in everyday setting or have too few (i.e. only 1 or 2 parts) or too many parts (composite scenes). Specifically, we kept 100 entities that are common everyday things that a child would be familiar with, with a mix of natural and man-made things. This annotation task involves answering the following question for each item in the list:
``Do you imagine this is something that most people would have seen in their everyday lives?''

We recognize there could be many variants of a single everyday entity e.g. different types of coffee makers. To narrow down the possibilities, the authors picked a diagram for each everyday thing via web search and carefully annotated a parts list for each of them to guide the level of granularity we are looking for. In some cases, the entity name was qualified to disambiguate further e.g. ``digital clinical thermometer'' instead of just ``thermometer.''

\subsection{Mental model annotations}

We ask crowdworkers to draw 
sketches of everyday things covering spatial relations, connectivity, and functional dependencies between parts (Table \ref{fig:4dims}).
To encourage the format of the mental model graphs to be more standardized across annotators, we ask that the nodes (in circles) mainly contain labels from the ``Parts list'' provided. However, to collect mental models that are most natural to the workers, they were also told that they can ignore parts in the ``Parts list'' if they seem unimportant, or add extra parts that seem important. We also specified for edges to be labeled with the relations shown in Table \ref{fig:4dims}.\footnote{For ease of annotation, they do not need to repeat annotations that mean the same thing. e.g. if they annotated ($x$, above, $y$), they do not need to annotate ($y$, below, $x$) again. We automatically generate these in our data post-processing.}

Given the name of an everyday thing, list of parts, and example diagram, 3 crowdworkers were recruited to sketch mental models for each everyday thing.\footnote{More details can be found in Appendix \ref{appendix:turk-task}.} Figure \ref{fig:example_mm} shows examples of such sketches. According to \citet{Norman2013TheDO},
mapping that takes advantage of spatial analogies leads to immediate understanding and is more natural. Sketching out such a graph allows workers more flexibility in taking advantage of spatial analogies between the actual entity and the sketch (see flashlight example in Figure \ref{fig:example_mm}). Therefore, we hypothesize that drawing a graph would be easier or more natural for crowdworkers than typing a list of relations.\footnote{Later these sketches are transcribed into (x, r, y) tuples.}

\subsection{Statistics}
\label{sec:dataset_stats}
\dataset{} consists of 100 everyday things ranging from devices like coffee maker, space heater to natural entities like tree and butterfly with number of parts (provided as a seed list to crowdworkers) ranging from 3-14.  
We collected 3 mental models per everyday thing. 
We take the \mms{} annotated by crowdworkers to be correct but not complete.
I.e., they may include only those relations that they think are salient for the everyday thing, and also omit the ones that can be easily inferred from what they have annotated e.g., when (trunk is $above$ the roots) is annotated, (roots are $below$ the trunk) can be omitted (Figure \ref{fig:example_mm}, tree example). For each everyday thing's mental model annotation, with the relation tuples annotated, we automatically add relations that are implied via enrichment based on 4 types of constraints (symmetric, asymmetric, inverse, and transitive). The inferred relations include both relations that are labeled True (e.g. A above B being True implies that B below A is True) and relations that are labeled False (e.g. A above B being True implies B above A is False). This gives a total of 11.7K gold relation tuples (6894 with ``True'' as gold labels and 4826 with ``False'' as gold labels). Table \ref{table:dataset_stats} provides additional dataset statistics. Appendix \ref{appendix:parrot-plus-plus} discusses the unanimity and diversity of mental models for these everyday things.

\section{Measuring and Improving Parts Mental Models} \label{sec:proposed-method}
Our proposed approach, \method,\footnote{First obtain the output of ``stochastic \textit{\underline{parrot}s},'' \citep{bender-parrots-21} then apply \underline{con}straints to reason on top of the output.} 
comprises two main components.\footnote{See Appendix \ref{appendix:illustration-parrot-con} Figure \ref{fig:constrained_plm} for an illustration.} The first component ``Probing a Pre-trained Language Model'' sends an exhaustive list of relation queries to a LM querying for every relation between each pair of parts (e.g. all relationships between egg white, yolk, shell, shell membrane and air cell). This gives us a large set of candidate relation tuples along with the model's confidence in each of them. Incorrect relation predictions can result in inconsistencies in the mental model. E.g, ``egg white both surrounds and is surrounded by the egg shell.''  The second component ``constraint reasoning'' then applies a constraint satisfaction layer on top of these raw predictions to choose a subset of these relation tuples that are maximally probable and minimally conflicting with each other. Note that \method{} is a zero-shot approach, where both probing LMs and constraint reasoning steps do not require any task-specific fine-tuning or re-training.

\subsection{Probing a Pre-trained Language Model}
We use the following pre-trained language models for our study:  GPT-3 \cite{NEURIPS2020_1457c0d6} and Macaw\footnote{A SOTA T5-11B based question-answering system \edit{that outperforms GPT-3 on some QA tasks}.} \cite{tafjord2021general}.
We probe them using True/False questions of type:
``Judge whether this statement is true or false: In an <everyday thing>, <part1 relation part2>.'' For each query $q$, we record an answer $a \in \{True, False\}$, and the model's beliefs about the likelihood of the relation being ``True'' as 
\begin{equation*}
\frac{p (True | q)} 
{p (True | q) + p (False | q)} .
\end{equation*}

\subsection{Constraint Reasoning } \label{sec:constrained_reasoning}
We observed a significant amount of inconsistency in raw predictions from these LMs by considering the following constraints:

\begin{itemize}
\item  \textbf{Symmetric relations:}  This constraint ensures symmetric 
relations like ``directly connected to'' and ``next to'' hold both ways.
\\i.e. 
$x$ rln $y$ $\leftrightarrow$ $y$ rln $x$

\item \textbf{Asymmetric relations:} For asymmetric relations like part of, has part, inside,
contains, in front of, behind, above, below,
surrounds, surrounded by, requires, required by, this constraint makes sure that both ``$x$ rln $y$'' and ``$y$ rln $x$'' cannot be true at the same time.  
\\i.e.
$\neg$($x$ rln $y$) $\lor$ $\neg$($y$ rln $x$)

\item \textbf{Inverse relations:} For a set of inverse relations e.g. above vs below, this constraint makes sure that ($x$ above $y$) and ($y$ below $x$) have the same truth value. 
\\i.e. 
$x$ rln $y$ $\leftrightarrow$ $y$ inverse(rln) $x$

\item \textbf{Transitive relations:} For relations like inside,
contains, in front of, behind, above, below,
surrounds, surrounded by, this constraint  will impose transitivity. 
\\i.e. 
$x$ rln $y$ $\wedge$ $y$ rln $z$ $\rightarrow$ $x$ rln $z$
\end{itemize}

\begin{table*}[ht!]
\centering
{\small
\setlength{\tabcolsep}{2pt}	
\begin{tabular}{l|c|ccccc|c} 
&    & 
                 \multicolumn{5}{c}{\makecell{\textbf{\%Conditional Violation (lower is better)}}}
                \\
                 & \makecell{\%True \\tuples} & 
                \makecell{Symmetric\\relations} &
                \makecell{Asymmetric\\relations} &
                \makecell{Inverse\\relations} &
                \makecell{Transitive \\ relations} & 
                \makecell{Avg.\\(macro)} &
                \makecell{Avg.\\(micro)}
                \\
\hline
\makecell{GPT-3\\(text-davinci\\-003)}      & 12.64 &
                                          \makecell{66.37\\(1,987/2,994)}  & \makecell{23.01\\(4,699/20,422)}   & \makecell{71.14\\(13,869/19,495)}  & \makecell{32.18\\(6,550/20,354)} &
                                          48.17 &  \makecell{42.84 \\ (27,105/63,265)}\\
\hline
Macaw-11B                                & 57.77  & 
                                          \makecell{29.98\\(3,089/10,305)}  & \makecell{64.97\\(42,170/64,910)}   & \makecell{33.63\\(21,642/64,361)}  & \makecell{10.08\\(44,121/437,746)} & 
                                          34.66& \makecell{19.23 \\ (111,022/577,322)}\\
\hline
\end{tabular}
}
\caption{\Mms{} constructed by LMs are significantly inconsistent with respect to their own predictions, violating basic commonsense constraints.
In brackets, we indicate (\# violations) / (\# constraints fired). \label{table:results-consistecy} }
\end{table*}


In this step, we try to resolve inconsistencies in LMs' raw predictions by solving a MaxSAT constraint satisfaction problem where each ($x$, relation, $y$) tuple is represented as a variable with confidence value from the LM used as its weight (soft clause). We introduce 4 types of hard constraints (listed above) between these variables as hard clauses and any constraint violation results in an extremely high penalty. Given a WCNF formula with these, a weighted MaxSAT solver tries to find an optimal assignment of truth values to relation tuples that maximizes the sum of weights of satisfied soft clauses and satisfies all the formula's hard clauses. 
We use the RC2 MaxSAT solver \cite{rc2} in PySAT \cite{imms-sat18}. 

\section{Results and Analysis}
\subsection{Evaluation Metrics}
We evaluate the \mms{} produced by the two LMs in terms of 
accuracy and consistency:

\noindent \textbf{Accuracy:} We compute the True/False accuracy of \mms{} based on the 11.7K gold relation tuples present in \dataset{}.

\noindent \textbf{Consistency:} Following \citet{kassner-etal-2021-beliefbank, mitchell2022enhancing}, we adapt the Conditional Violation ($\tau$) \cite{li2019logic} metric to measure inconsistency across the 4 types of constraints defined in Section \ref{sec:constrained_reasoning}. 
    For constraints $L(x) \rightarrow R(x)$ imposed on samples $x \in D$, where $D$ is the dataset, we calculate conditional violation as:
    \begin{equation*}
    \tau =
    \frac{
      \sum\limits_{x \in D} \left[ \bigvee\limits_{(L,R)} \neg{(L(x) \rightarrow R(x))} \right]}{\sum\limits_{x \in D} \left[ \bigvee\limits_{(L,R)} L(x) \right]} .
    \end{equation*}
    


\subsection{Results}

\noindent \textbf{Q1: How consistent are LMs when they answer questions about everyday things?}

\noindent We measure the consistency of \mms{} constructed by LMs based on 4 types of constraints described in Section \ref{sec:constrained_reasoning}. This measurement is purely based on LMs' predictions and is independent of relations in the gold mental models acquired for the everyday things. Table \ref{table:results-consistecy} shows that LMs contradict themselves (19-43\% conditional violation) when we ask them multiple questions about parts of the same everyday thing to probe for their \mm{}.
E.g., in Appendix \ref{appendix:illustration-parrot-con}, the LM believes that in an egg, ``yolk surrounds the shell'' and ``shell surrounds the yolk'' are both True. 
Table \ref{table:results-consistecy} also breaks down the LMs' inconsistency across 4 types of constraints. We observe that GPT-3 struggles with maintaining consistency for symmetric and inverse relations, whereas Macaw-11B finds it most challenging to satisfy constraints for asymmetric relations.


\vskip 0.8cm

\noindent \textbf{Q2: Do language models have accurate mental models of everyday things?}

\noindent Next, we investigate how accurate are these \mms{} when compared to gold mental models in our \dataset{} dataset. Table \ref{table:results-accuracy} shows that such queries pertaining to parts of everyday things are challenging for even SOTA models, with an average accuracy of 54-59\%. This is barely better than the majority class baseline at 59\% and random chance at 50\%. 
\begin{table}[h!]
\centering
{\small
\setlength{\tabcolsep}{3pt}	
\begin{tabular}{l|l|c|cc} 

                &  \makecell{\# params} & \makecell{Base \\LM (\%)}  & \makecell{\method \\ (\%)} & \makecell{Improve \\(\%)}\\
\hline
\makecell{GPT-3 (text-\\davinci-003)}   & 175B   & 53.83   & 70.26 & 16.42 \\
\hline
Macaw-11B                      &   11B       & 59.45    & 79.28 & 19.84\\
\hline

\end{tabular}
}
\caption{Comparing the accuracy of \mms{} before and after constraint reasoning on \dataset{} dataset. 
\label{table:results-accuracy} }
\end{table}

The LMs' low performance shows that \dataset{} is a challenging dataset, which is expected given the fact that this dataset queries for commonsense knowledge about everyday things (e.g. spatial relationship between parts of a device) that are often omitted in text, and hence less likely seen during pre-training. Further, by construction, our queries minimally differ e.g. for relations between parts of a tree, the edit distance between a statement with true relation ``the leaves are above the roots''  and false relation ``the leaves are below the roots'' is just 1 word. This makes our task even more challenging as the models need to understand the semantics of relational phrases to give the correct answer.

\vskip 0.8cm

\noindent \textbf{Q3: Does \method{}, our proposed constraint reasoning approach, help create more accurate mental models?}

\noindent Our proposed approach, \method{}, utilizes the inherent inconsistency in LMs' raw predictions to self-correct their own \mms{}. 
It finds an  optimal assignment of truth values to relation tuples that accounts for both the model's original beliefs (about the likelihood of each relation statement being True or False), and the 4 types of commonsense constraints imposed.
By imposing the commonsense constraints as hard constraints,
our proposed method produces perfectly consistent mental models for all LMs with respect to the imposed constraints
i.e. \% conditional violation becomes 0 for all columns in Table \ref{table:results-consistecy}.
Using these basic commonsense constraints, \method{} improves \mm{} accuracy significantly by 16-20\% on \dataset{} (Table \ref{table:results-accuracy}).


    \begin{figure}[t]
    \small
    \centering
         \includegraphics[width=1.01\columnwidth]{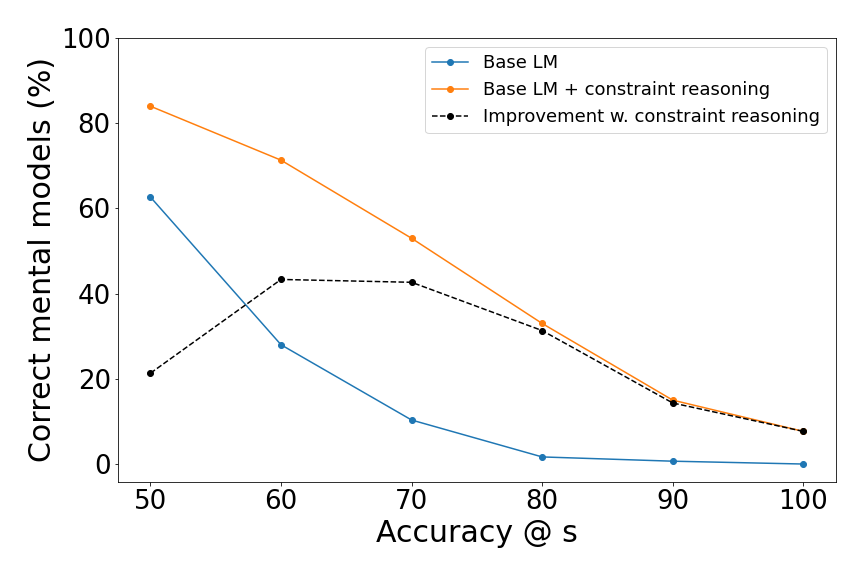} 
         \\
    (a) GPT-3
         \includegraphics[width=1.01\columnwidth]{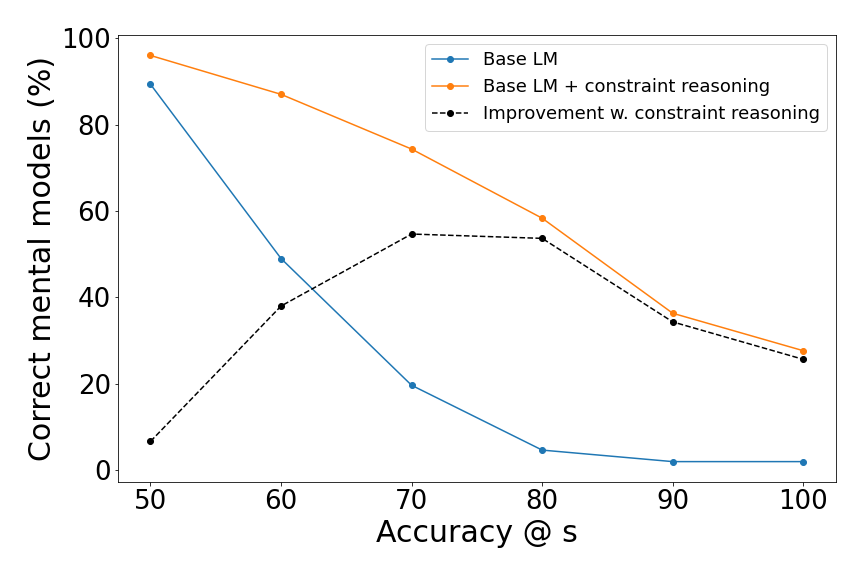}
         \\
         (b) Macaw-11B
    \caption{Percentage of correct mental models vs accuracy@$s$ shows that for both GPT-3 and Macaw-11B, there is a higher percentage of correct mental models after constraint reasoning (orange) as compared to raw LM predictions (blue), no matter the threshold for considering a mental model to be correct is lower or higher.
    For improvements from constraint reasoning (black), we observe the highest increase in percentage of mental models that are at least 60-80\% accurate. 
    \label{fig:lm-threshold-accuracy-macaw-gpt3}}
    \end{figure}

\begin{figure}[t]
\small
\centering
    \subfloat[\centering GPT-3]{\includegraphics[width=0.925\columnwidth]{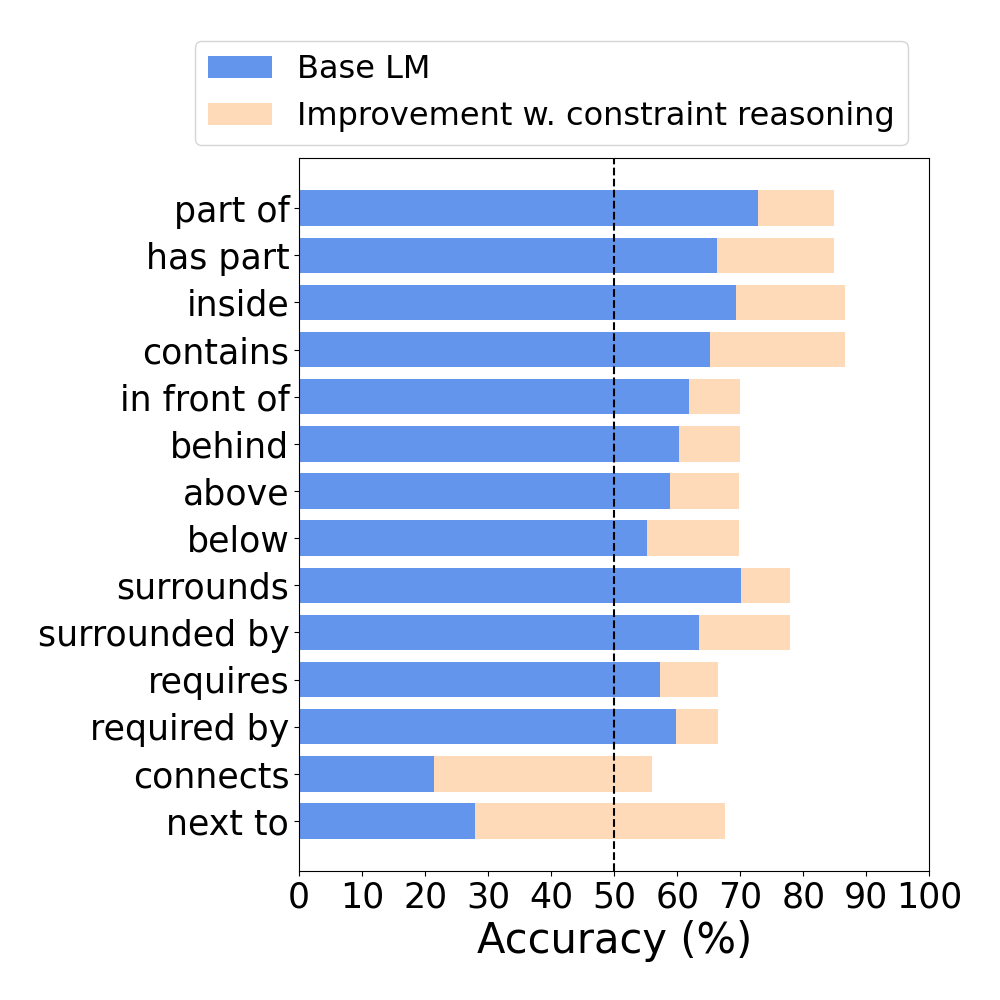}}\\
    \subfloat[\centering Macaw-11B]{\includegraphics[width=0.925\columnwidth]{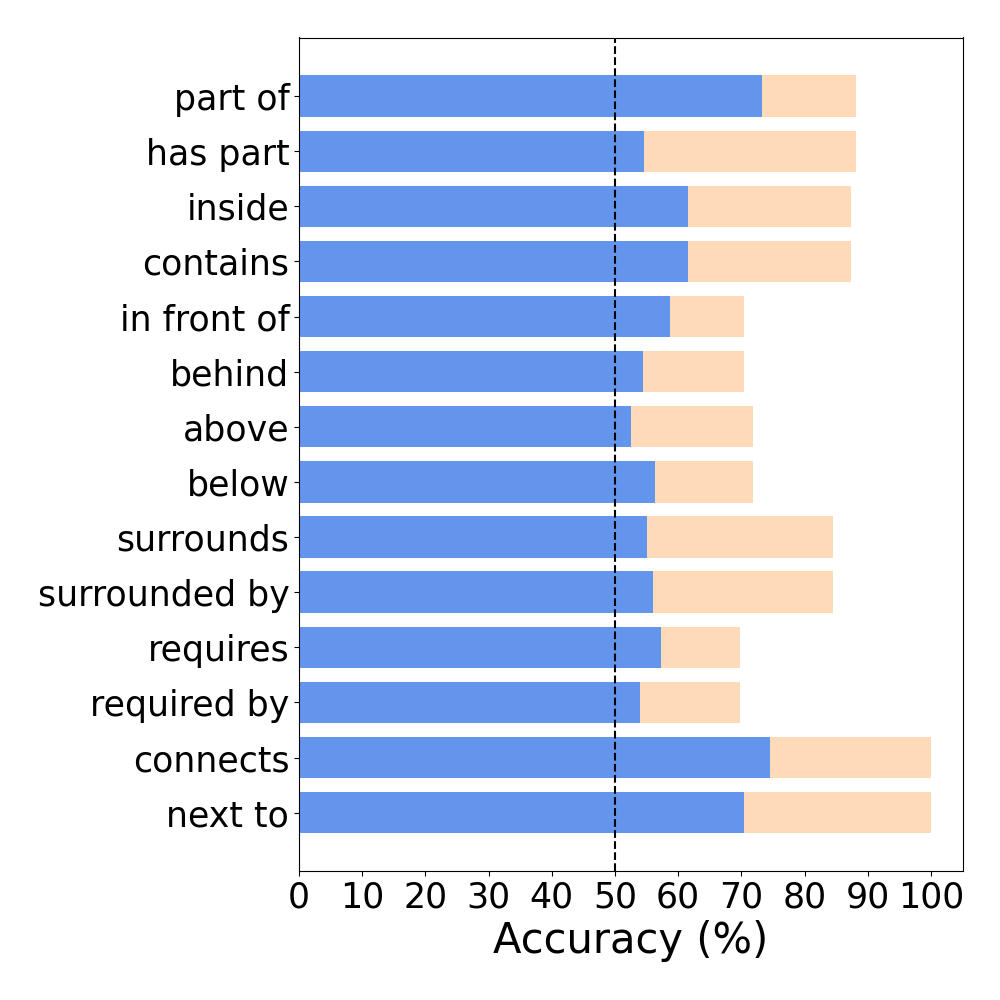}}
\caption{Accuracy of base LM and improvement achieved through constraint reasoning on different relations in \dataset{} dataset.
\label{fig:accuracy-by-relation}}
\end{figure}

\subsection{Further analysis}
\label{sec:analysis}
\noindent \textbf{Most effective range} \quad We analyze what is the quality range of mental models that \method{} is most effective on. 
We quantify the quality of \mms{} by defining accuracy@$s$, a metric that says a mental model is correct if the proportion of correct relations is at least $s$\%. We then plot the percentage of mental models (out of 300) that are correct vs accuracy@$s$ for different values of $s$, where $s \in \{50, 60, 70, 80, 90 , 100\}$.
Figure \ref{fig:lm-threshold-accuracy-macaw-gpt3} shows that \method{} not only effectively increases the percentage of mental models that are approximately correct ($s$ = 50, 60) but also the percentage of mental models that are (almost) totally correct ($s$ = 90, 100). The improvements with constraint reasoning are even more prominent when it comes to increasing the percentage of mental models that are at least 60-80\% accurate. This is likely attributed to the improvement in mental models that have enough signals from LMs’ raw predictions and also enough margin to improve. 

\vskip 0.3cm
\noindent \textbf{ Accuracy of \mms{} per relation} \quad Figure \ref{fig:accuracy-by-relation} shows that the base LMs are more accurate in predictions for queries containing relationships like `part of' which is more likely to be stated in text than spatial relations like `above', `below', and `behind' which are lower-level physical details often not mentioned in text. 
Different models also differ in which relationships they perform better on: e.g. GPT-3 performs poorly on bi-directional relations like `connects' and `next to', with accuracy way below chance level, while Macaw-11B achieves around 70\% accuracy for queries involving these relations. 

    \begin{figure}[t]
    \small
    \centering
         \includegraphics[width=\columnwidth]{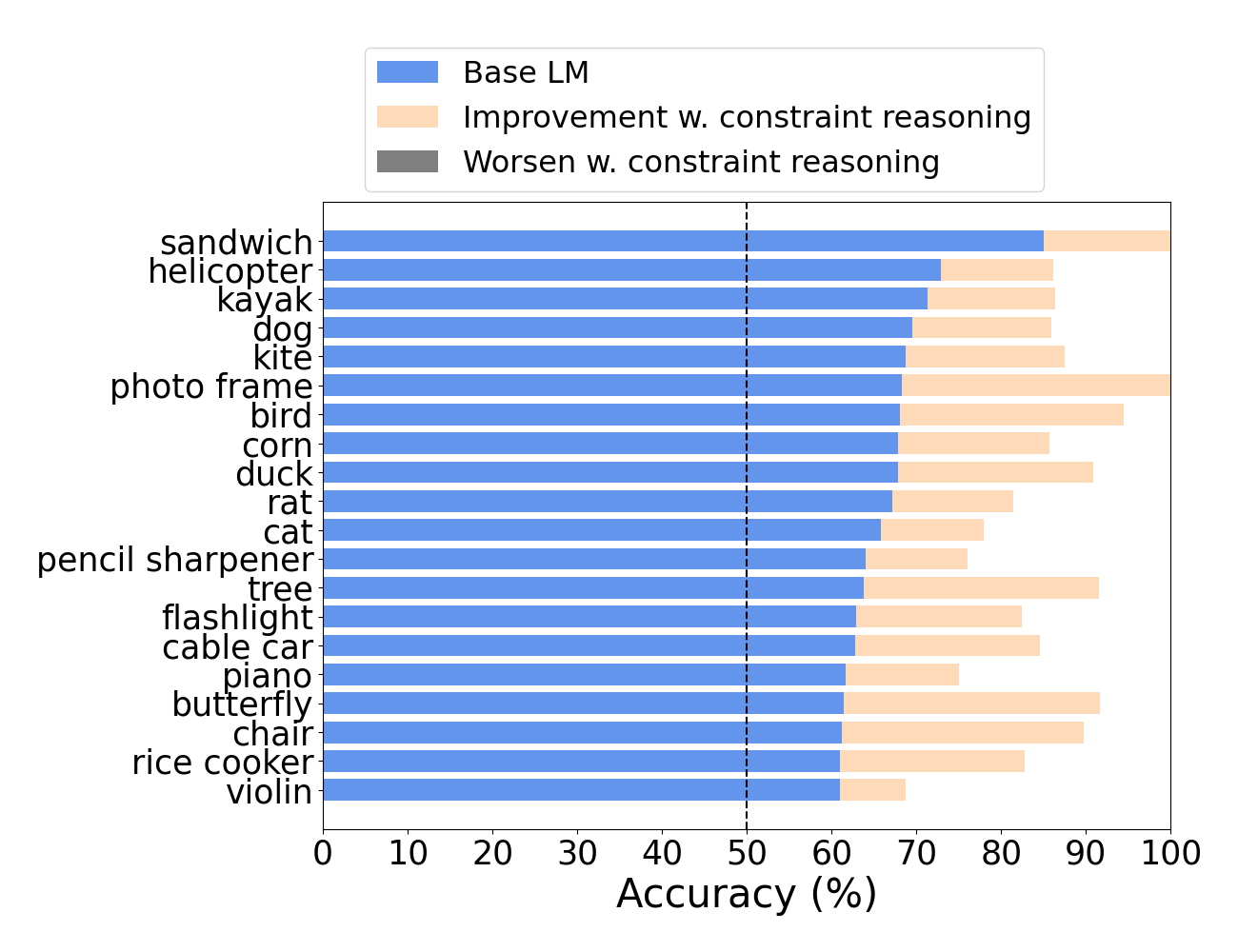} 
    (a) GPT-3
         \includegraphics[width=\columnwidth]{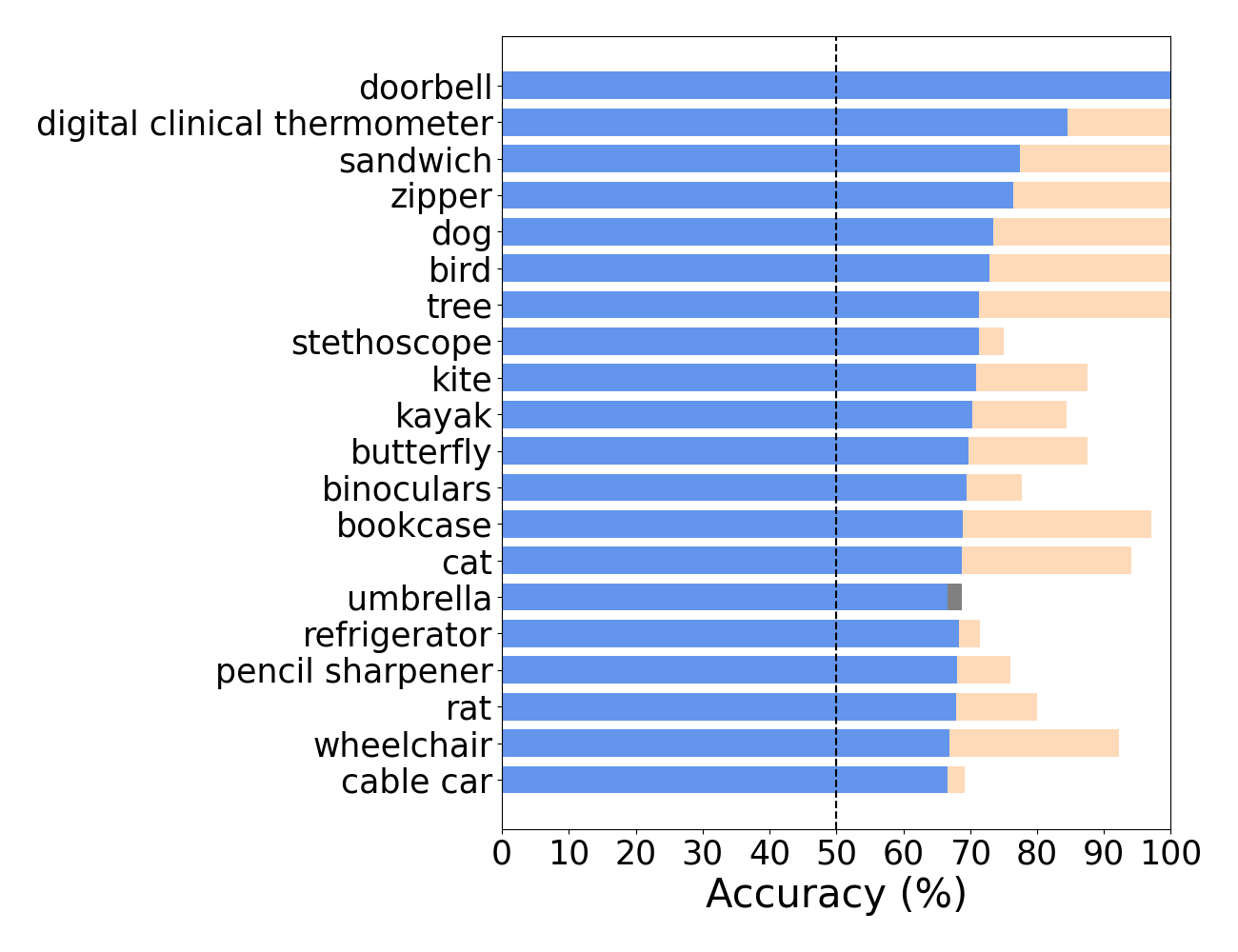} 
         (b) Macaw-11B
    \caption{20 everyday things that each model achieved \textbf{best performance} on, based on models' raw predictions (i.e. Base LM). In almost all cases, constraint reasoning boosts the accuracy of the \mms{} produced by the base LM, pushing it even closer to 100\%.
    \label{fig:accuracy-by-et-best}}
    \end{figure}
    \begin{figure}[th]
    \small
    \centering
         \includegraphics[width=\columnwidth]{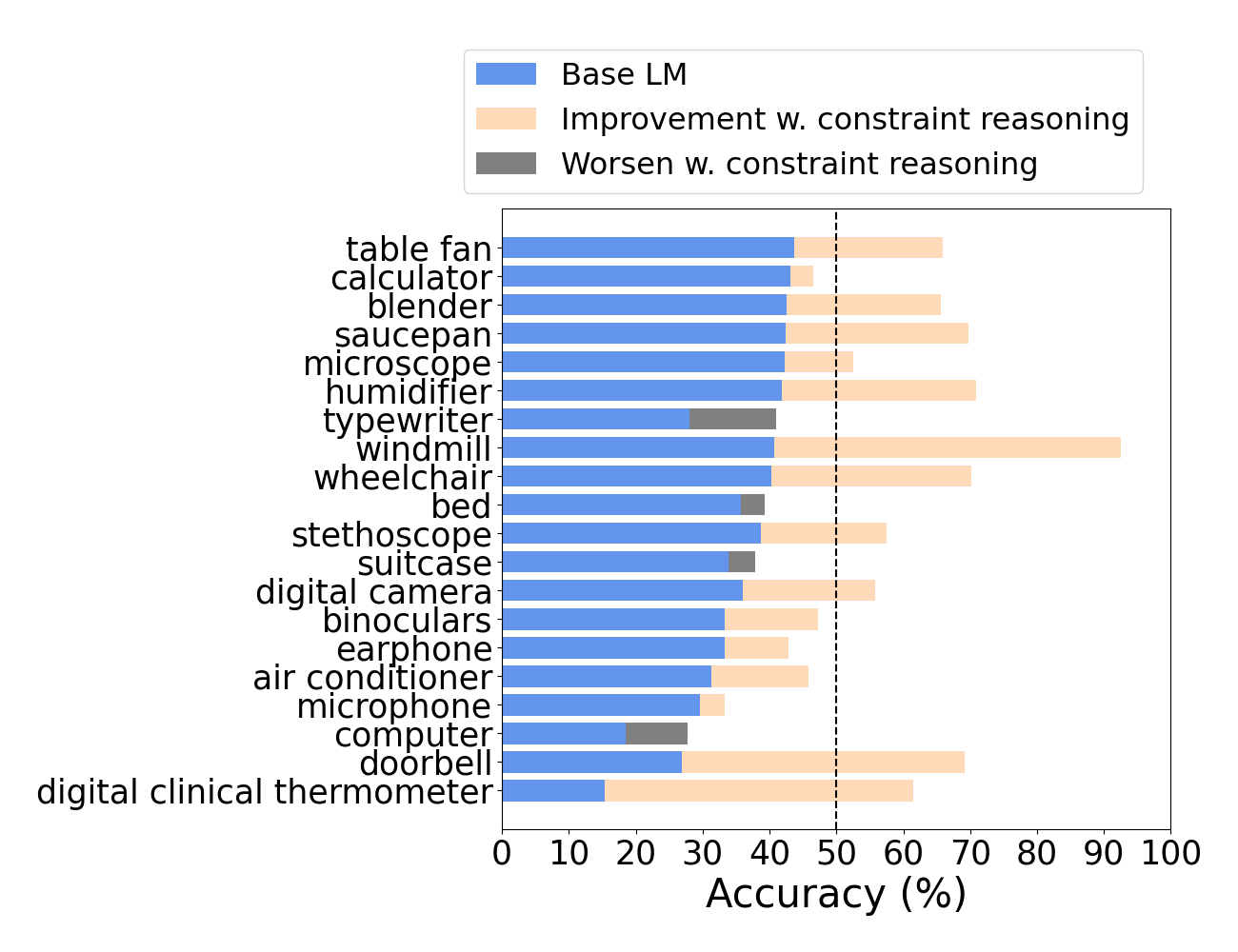} 
    (a) GPT-3
         \includegraphics[width=\columnwidth]{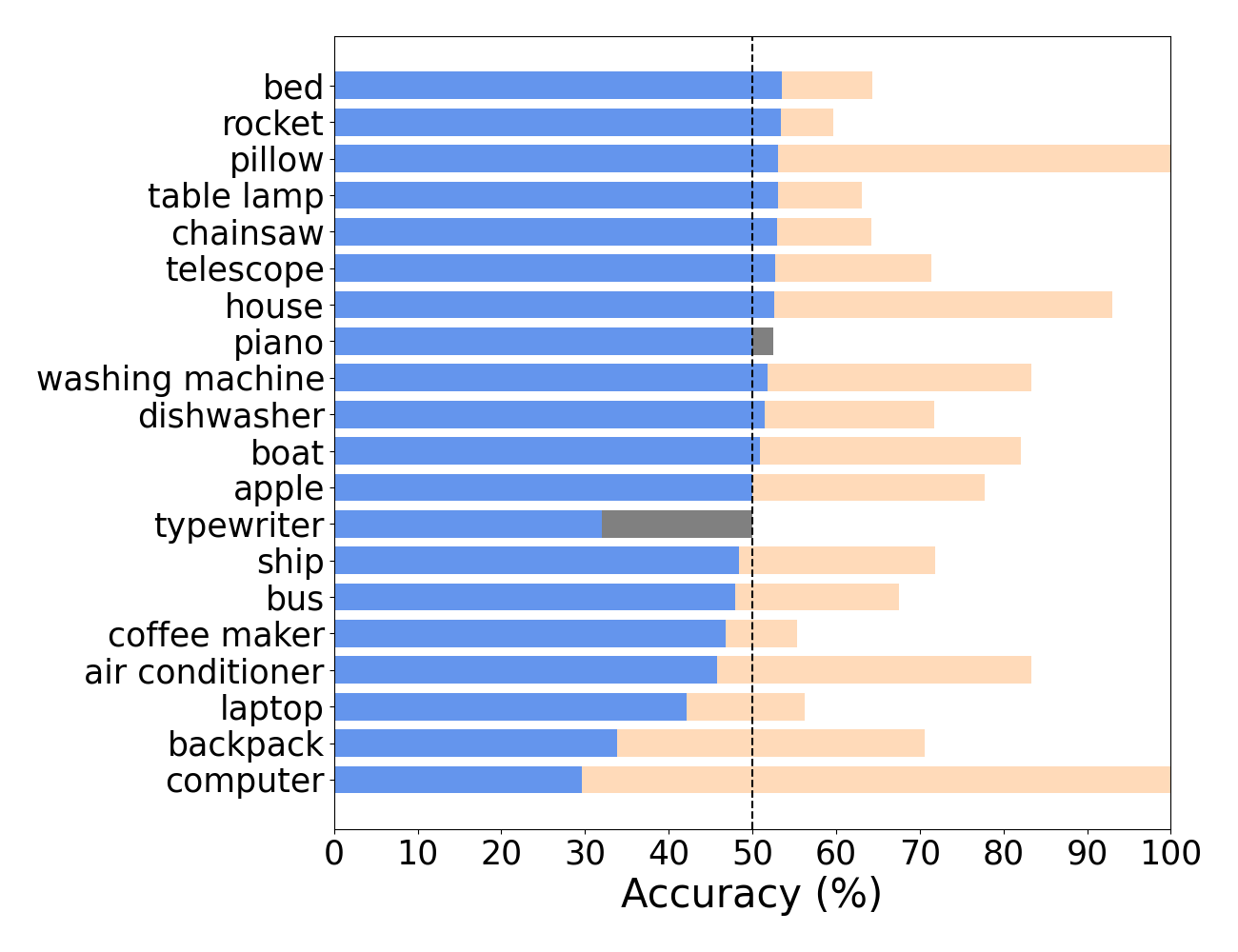} 
         (b) Macaw-11B
    \caption{20 everyday things that each model achieved \textbf{worst performance} on, based on models' raw predictions (i.e. Base LM). In many of these cases, the accuracy of the \mms{} produced by the base LM is at around or below chance level and constraint reasoning boosts accuracy to beyond 50\%.
    \label{fig:accuracy-by-et-worse}}
    \end{figure}

\vskip 0.3cm
\noindent \textbf{Success and failure across models per everyday thing} \quad LMs show both similarities and differences in what everyday things they have better mental models of. For each model, Figure \ref{fig:accuracy-by-et-best} shows the top 20 everyday things that the models performed \textit{best} on in terms of base LM accuracy. Both GPT-3 and Macaw-11B perform well on the following everyday things: sandwich, kayak, dog, kite, bird, rat, cat, pencil sharpener, tree, cable car, and butterfly. It is interesting to see that both models perform well on several natural living things like animals (e.g. dog, bird, rat, cat), insect (e.g. butterfly), and plant (e.g. tree). Figure \ref{fig:accuracy-by-et-worse} shows the top 20 everyday things that the models performed \textit{worst} on in terms of base LM accuracy. We observe that entities like typewriter, bed, air conditional, and computer are challenging for both models to form accurate mental models of. Although the models share some similarities in what everyday things they have better/worse mental models of, they also show differences, especially for man-made devices: e.g. GPT-3 does well but Macaw-11B performs poorly on forming an accurate \mm{} of piano; Macaw-11B does well, but GPT-3 performs poorly on devices like doorbell, digital clinical thermometer, and binoculars.



\section{Conclusion}
Do language models have coherent mental models of everyday things? To systematically study this question, we present a benchmark dataset, \dataset, consisting of 300 human-constructed mental models for 100 everyday objects, including over 2K parts and 11.7K relationships between these parts. Our experiments reveal that even SOTA LMs generally have poor mental models (inaccurate and violating basic commonsense constraints) of everyday things, thus providing insight into their apparent knowledge and behavior not previously explored. 
We apply constraint reasoning on top of base LM predictions to construct more coherent mental models.
Our method, \method{}, improves both accuracy (up to 20\% improvement) and consistency (up to 43\% improvement) of such \mms{}. This suggests a broader cognitive architecture
(LM + reasoner) for future systems, to construct more coherent mental models than using the LM alone.

\clearpage
\section*{Limitations}
Common everyday things change over the years. While we try to choose ones that are in children's vocabulary, over decades, devices evolve and humans change in which things they interact with more frequently, affecting which relationships would be more prominent in an average person's mental model. So the \mms{} in such a dataset may not stay constant over time (e.g. some entities may be less familiar and certain relations may be less salient to annotators of the future). It would be interesting to use our \dataset{} dataset as a point of comparison when studying mental models of everyday things in the future to reveal interesting insights on how humans' mental models of everyday things evolve over time.

Other important future directions include to explore how more coherent mental models can help in complex reasoning tasks about everyday things, combine these \mms{} with mental models along other dimensions e.g. \citet{gu-etal-2022-dream, gu-etal-2022-dreamflute}, as well as using our dataset of commonsense queries about everyday things as a source of follow-up questions for existing QA tasks e.g., PIQA \cite{Bisk_Zellers_Lebras_Gao_Choi_2020} and CSQA \cite{csqa}. 

This paper only focuses on relationships (spatial orientation, connectivity, and functional dependency) between parts of everyday things. However, our approach \method{} is easily extensible to other applications such as:
\begin{itemize}
    \item spatial relations in other domains e.g. for geographical distances, we can similarly impose constraints on inverse relations like \textit{closer} and \textit{further}
    \item temporal relations e.g. on a timeline, if event A occurred \textit{before} event B, then event B cannot have occurred \textit{before} event A (\textit{before} is asymmetric)
\end{itemize}
We leave the demonstration of the generalizability of our approach to future works.


\section*{Ethics Statement}
All annotators that participated in the data collection process have been anonymized.  The only personal information we collect is the worker IDs from Amazon
Mechanical Turk, which we will not release. No  personally identifiable information is contained in our dataset or otherwise released. We took great care to pay fair
wages, and were responsive to feedback and questions throughout the data collection process. This study involves the use of large-scale language models. We only use them to generate True/False answers to questions about parts of everyday things, therefore we do not foresee any substantial ethical issues with their use for research presented in this submission.

\section*{Acknowledgements}
\edit{We thank the anonymous ACL reviewers, as well
as Ernest Davis, Chris Callison-Burch and
members of the Aristo team at AI2 for their
valuable feedback on an earlier draft.}

\bibliography{anthology,custom}
\bibliographystyle{acl_natbib}

\appendix

\onecolumn
\section{Source of everyday things}
\label{appendix-source-of-etc}
We compiled a list of 100 everyday things from:

\begin{enumerate}
   \item Children’s books
    \begin{enumerate}
     \item My First Library series \edit{\citep{firstlibrary2018}}
     \item Now you know how it works \edit{\citep{nowyouknow2019}}
     \item My first 100 things that move \edit{\citep{firstthingsthatmove2018}}
   \end{enumerate}
   \item Vocabulary lists
    \begin{enumerate}
     \item Grade 1-5 vocabulary list \edit{\citep{basicspellinglist}}
     \item Select from all the nouns from an 8th-grade vocabulary list that were also under either “artifact” or “device” in WordNet \edit{\citep{miller-1994-wordnet}}
   \end{enumerate}
 \item Online web search
 \end{enumerate}

\section{Details on mental model annotation task} \label{appendix:turk-task}

\textbf{Mechanical Turk task instructions:} \\ \\
\includegraphics[width=0.95\columnwidth]{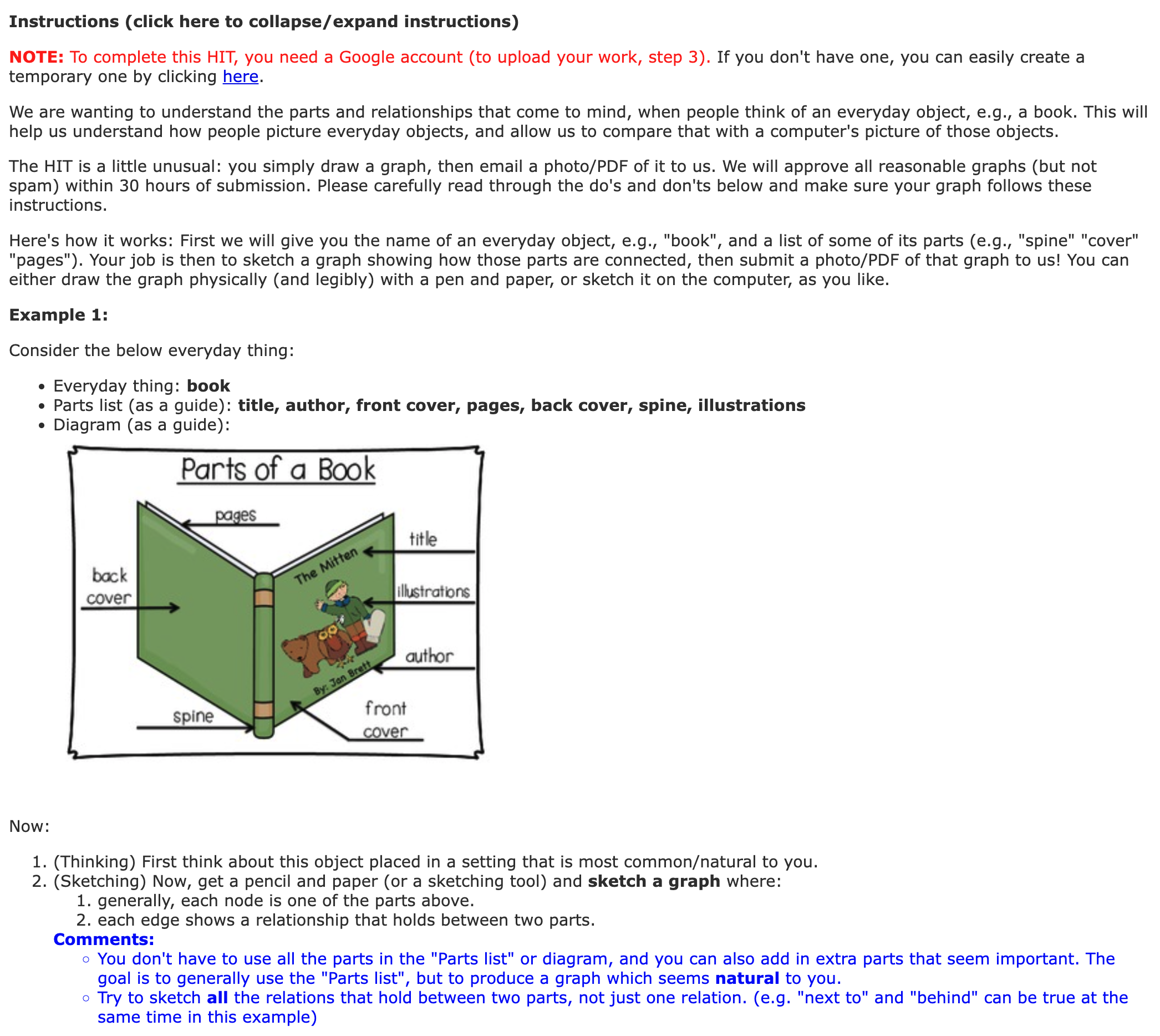}
\includegraphics[width=0.95\columnwidth]{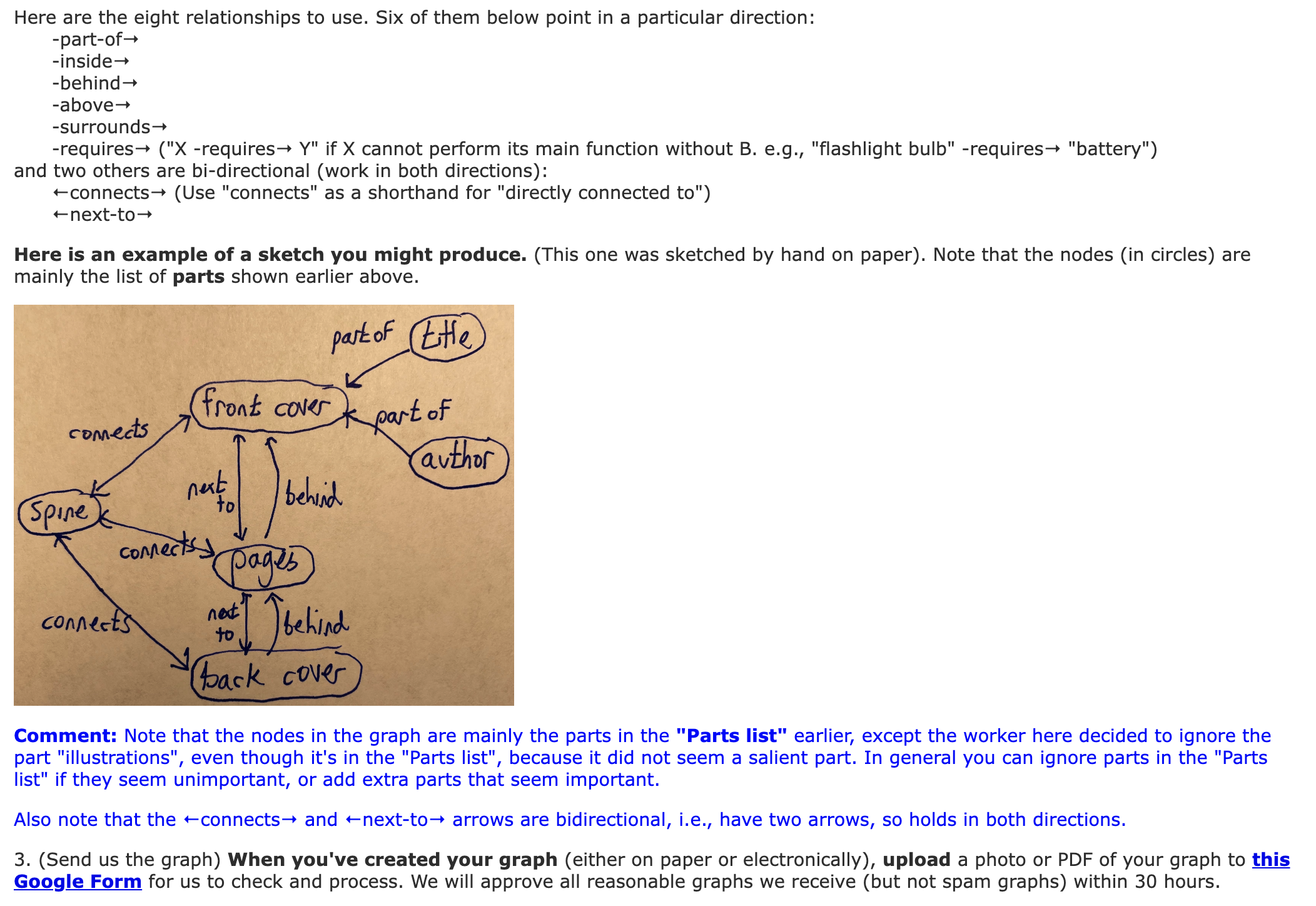}

Our participants were recruited on the Amazon Mechanical Turk platform. The workers met minimum qualification in AMT: 95\% approval rate. They were from US locations and rated at Amazon’s Masters Level. 
Workers were paid at a rate of $\approx$\$15/hr.

\section{Unanimity and diversity in \mms{}} \label{appendix:parrot-plus-plus}
People vary greatly in how they construct mental models, but the underlying reasoning is often structurally similar i.e. in accordance with commonsense constraints \citep{Halford1993ChildrensUT, Jonassen1996MentalMK}. In our \dataset{} dataset, similarly, contradictions amongst crowdworkers (e.g., for guitar, one worker annotated that the neck is part of the fingerboard, while another annotated that the fingerboard is part of the neck) are extremely rare. There are only 80 instances out of 11720 in total in our entire dataset (0.68\%) -- less than 1\%.

\begin{figure*}[h]
\centering
     \includegraphics[width=\textwidth]{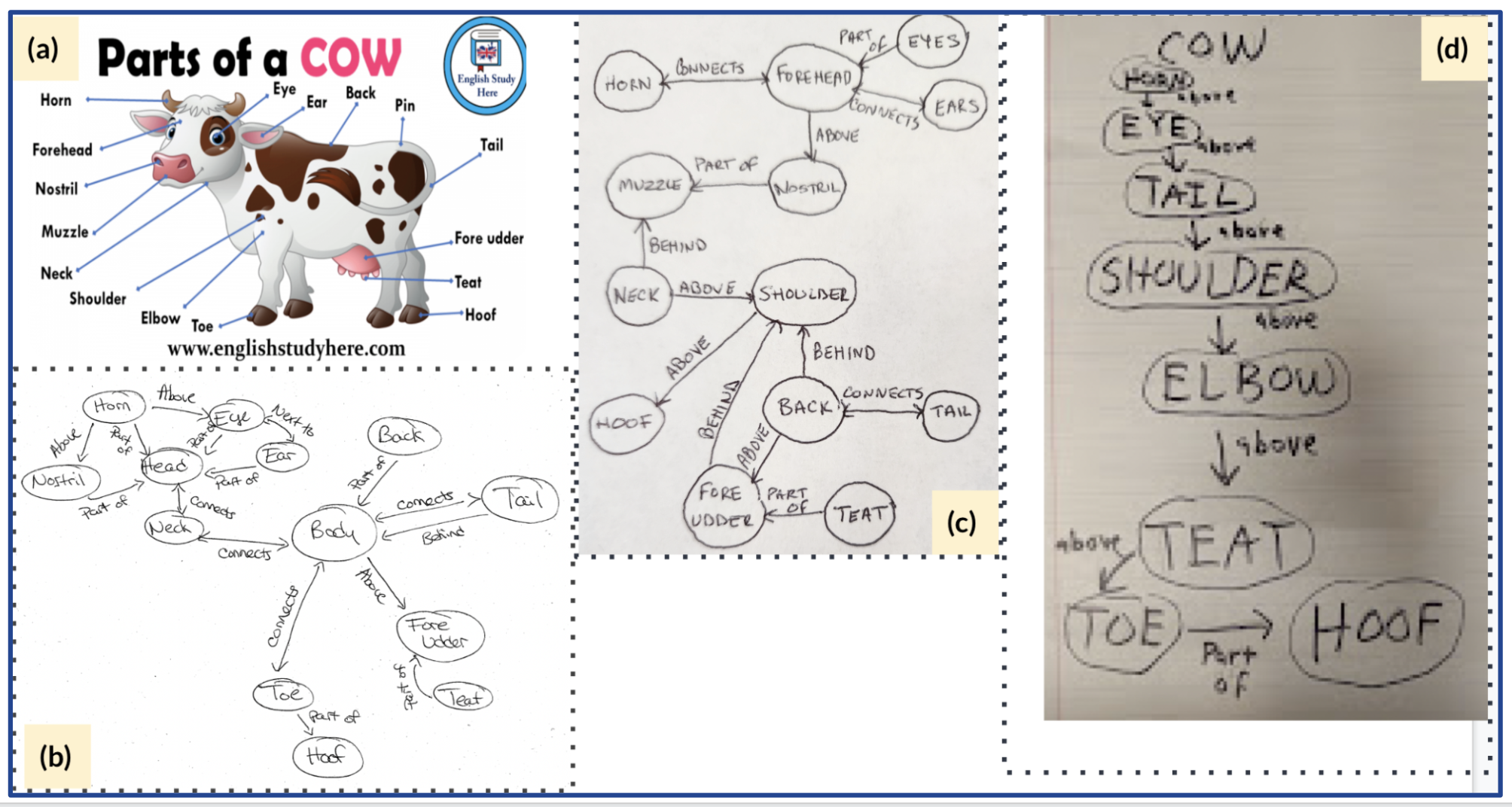}
\caption{ Example \mm{} annotations from \dataset{}: (a) we provide the crowdworkers a diagram of cow retrieved from the Web. (b), (c), (d) are \mm{} sketches by 3 different crowdworkers. Note that all 3  models are accurate but there is some divergence in terms of (1) part names: e.g., `head' vs `forehead' and (2) which relation  tuples they consider salient. Similar forms of diversity have been reported in \citet{ji2022abstract}, for instance, as part naming divergence and segmentation divergence.
\label{fig:mm_divergence}}
\end{figure*}

We also looked at relations overlapped across workers in our dataset to analyze if workers pay attention to similar or different aspects of everyday things. To do so, we gathered a set of (p1, rln, p2) relations that are common across all 3 annotators for each everyday thing. These relationships are ones that achieved full agreement across all the 3 assigned annotators for that everyday thing in terms of the spatial/connectivity/functional relationship annotated and the parts involved. Together, we refer to this set as the \dataset{}++ dataset. Table \ref{fig:common-relations} summarizes the number of such high-agreement relationships for each everyday thing.
Everyday things with few or no high-agreement relationships (refer Figure \ref{fig:mm_divergence} for an example) imply higher diversity among annotators in terms of which spatial/connectivity/functional relationship and what parts they decided to include in their annotations. There are a total of 508 overlapped relations in \dataset{}++, out of the 11720 in \dataset{}, suggesting that attention is often paid to different aspects of everyday things.

\begin{table} [h]
\centering
{\small
\setlength{\tabcolsep}{2pt}	
\begin{tabular}{c|l} 
\makecell{\# full-agreem.\\relations} & Everyday thing(s) \\ \hline
36 & coffee maker, fish\\ \hline
28 & rabbit \\ \hline
18 & deer \\ \hline
16 & egg, electric stove, tree \\ \hline
14 & ink pen \\ \hline
12 & \makecell[l]{laptop, sandwich, rice cooker, airplane, table} \\ \hline
10 & fire extinguisher, bird \\ \hline
8 & \makecell[l]{elevator, flashlight, stroller, dishwasher, kayak, ship, teapot, telescope,\\ corn, hot air balloon, microwave} \\ \hline
6 & \makecell[l]{wheelchair, barbeque grill, kite, microphone, computer, duck, helicopter} \\ \hline
4 & \makecell[l]{pillow, truck, washing machine, door, hair dryer, rocket, screw, toaster, \\butterfly, chair, knife, photo frame, shoe, baby bottle, bed, bird cage, \\car, chainsaw, electric tea kettle, humidifier, piano} \\ \hline
2 & \makecell[l]{binoculars, digital camera, zipper, apple, digital clinical thermometer, earphone, flower, \\windmill, backpack, dog, doorbell, lightbulb, bat, cat, umbrella, stethoscope, tent} \\ \hline
0 & \makecell[l]{air conditioner, bicycle, blender, boat, glider, guitar, house, pencil sharpener, \\table fan, dryer, pencil, suitcase, telephone, microscope, refrigerator, space \\heater, typewriter, violin, wall clock, window, bookcase, bus, cable car, calculator, \\saucepan, train, cow, rat, table lamp} \\ 
\hline
\end{tabular}
}
\caption{Number of relationships that achieved full agreement across all the 3 assigned annotators for each everyday thing. Higher number of such relations indicates more unanimous parts mental model annotations, whereas lower number reflects more diversity. 
\label{fig:common-relations} }
\end{table}

In Table \ref{table:results-accuracy-overlap-set}, we present accuracy on \dataset{}++, revealing similar results for relationships that achieved full agreement across all assigned annotators.
Using basic commonsense constraints, \method{} improves \mm{} accuracy significantly by 16-22\% on \dataset{}++. 
These trends are similar to that obtained for \dataset{}, illustrating that the results hold across all gold-standard parts relations, regardless of whether they are more unanimous or diverse across annotators.

    \begin{table} [h]
\centering
{\small
\setlength{\tabcolsep}{3pt}	
\begin{tabular}{l|l|c|cc} 

                &  \makecell{\# params} & \makecell{Base \\LM (\%)}  & \makecell{\method \\ (\%)} & \makecell{Improve \\(\%)}\\
\hline
\makecell{GPT-3 (text-\\davinci-003)}   & 175B   & 55.51   & 71.13 & 15.62 \\
\hline
Macaw-11B                      &   11B       & 60.04    & 82.41 & 22.38\\
\hline

\end{tabular}
}
\caption{Comparing the accuracy of \mms{} before and after constraint reasoning on \dataset{}++ dataset.\label{table:results-accuracy-overlap-set} }
\end{table}
\clearpage

\section{Pictorial illustration of \method{}} \label{appendix:illustration-parrot-con}
\begin{figure*}[h]
\centering
     \includegraphics[width=\textwidth]{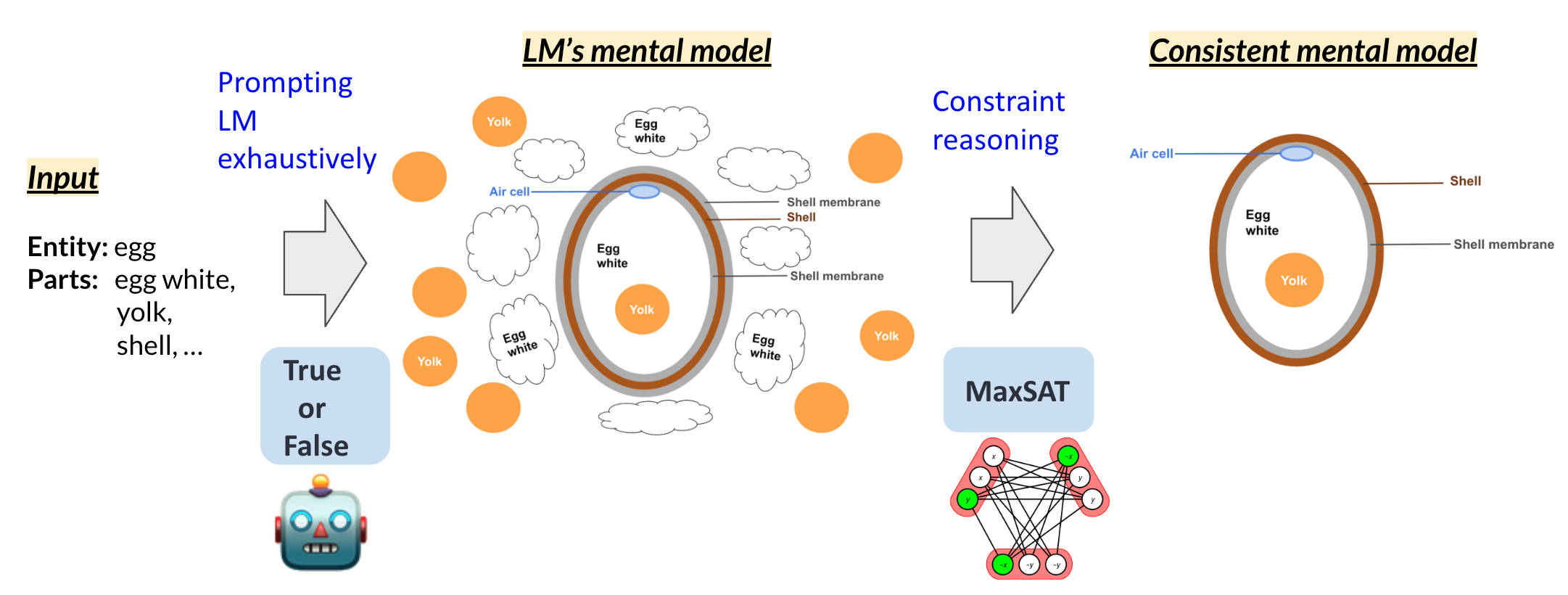}
\caption{When asked about relationships between parts of an everyday thing, LMs can produce inconsistent relations. E.g., GPT-3 believes that in an egg, ``yolk surrounds the shell'' and ``shell surrounds the yolk'' are both True. Our proposed neuro-symbolic method, \method, applies constraint reasoning over raw LM predictions to produce more accurate and consistent mental models of everyday things. 
\label{fig:constrained_plm}}
\end{figure*}

Our proposed approach, \method, is illustrated in Figure \ref{fig:constrained_plm} with an example everyday entity ``egg''. 

\section{Accuracy on different everyday things}
\begin{figure*}[]
\centering
\begin{subfigure}{0.49\textwidth}
  \centering
  \includegraphics[width=\linewidth]{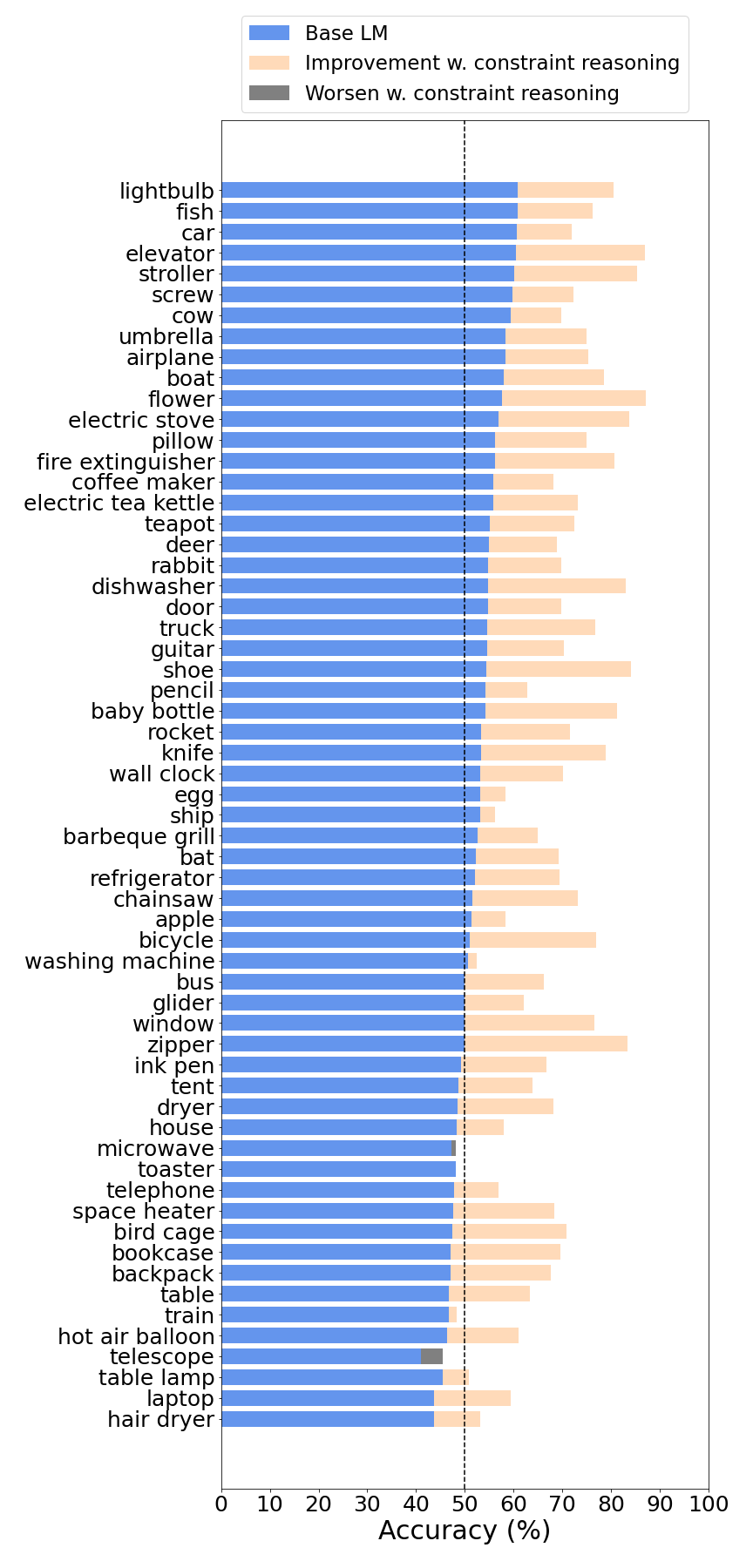}
  \caption{GPT-3}
  \label{fig:sub1}
\end{subfigure}%
\begin{subfigure}{0.49\textwidth}
  \centering
  \includegraphics[width=\linewidth]{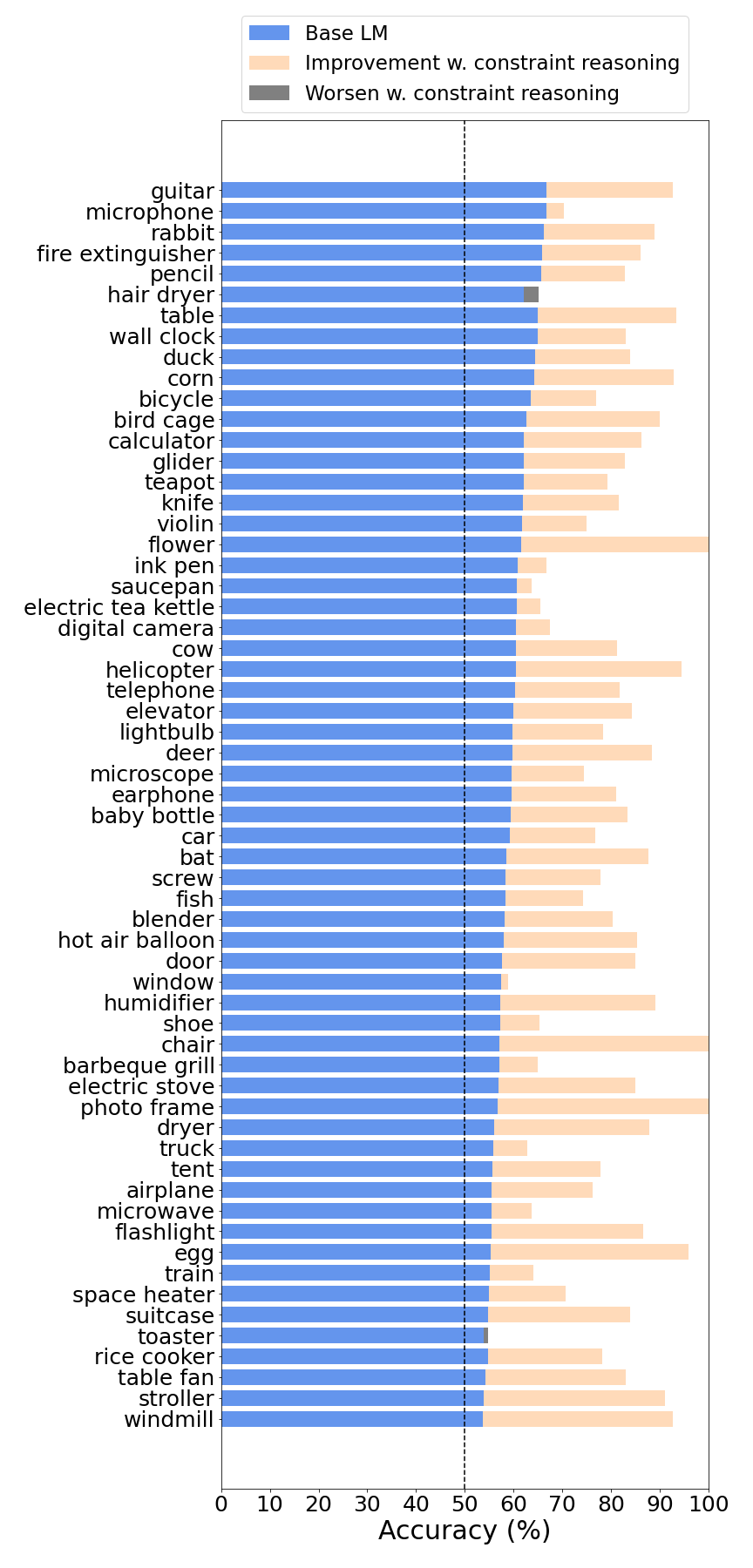}
  \caption{Macaw-11B}
  \label{fig:sub2}
\end{subfigure}
\caption{Performance on other everyday things. Accuracy of base LM and improvement achieved through constraint reasoning on different everyday things in our dataset.}
\label{fig:other-et}
\end{figure*}

Table \ref{fig:example-prompt-responses} gives example prompts and GPT-3's responses (includes both correct and incorrect) for entity ``tree''.
Top 20 and bottom 20 everyday things that each model achieved best and worst performance on are shown in Figures \ref{fig:accuracy-by-et-best} and \ref{fig:accuracy-by-et-worse} respectively. Further, Figure \ref{fig:other-et} demonstrates everyday things with 21st to 80th ranking in terms of the base LM accuracy.
\begin{table} [h]
\centering
{\small
\setlength{\tabcolsep}{2pt}	
\begin{tabular}{c|c|l} 
\makecell{Model} & Prompt & Model's Answer\\ \hline

GPT-3 & \makecell{Judge whether this statement is true or false: \\In a tree, twig is directly connected to the branches.} & True \textcolor{green}{(correct)}\\
\hline
GPT-3 & \makecell{Judge whether this statement is true or false: \\In a tree, trunk is above the roots.} & False \textcolor{red}{(incorrect)}\\
\hline
GPT-3 & \makecell{Judge whether this statement is true or false: \\In a tree, roots are surrounded by the trunk.} & True \textcolor{red}{(incorrect)}\\
\hline
GPT-3 & \makecell{Judge whether this statement is true or false: \\In a tree, trunk is below the roots.} & False \textcolor{green}{(correct)}\\
\hline
\end{tabular}
}
\caption{Example prompts and GPT-3's responses for an everyday entity ``tree''.
\label{fig:example-prompt-responses} }
\end{table}

\section{Use of models for inference}


For all experiments in this paper we used existing models/toolkits without any re-training or fine-tuning.
We used GPT-3  text-davinci-003  and Macaw (T5-11B based) as representative LMs for our experiments. 
To probe GPT-3  text-davinci-003, we used their web API which took around 30 to 60 msec per relation tuple (one T/F question).
To probe Macaw, we used two 48GB GPUs and it takes around 10.4 msec per relation tuple.
We also run a MaxSAT solver for each everyday entity's parts mental model. To solve a constraint satisfaction problem per parts mental model takes a few msec up to around 3 minutes depending on the WCNF formula involved. \\ 

\section{On the use of our dataset and code}
\edit{We have made all data and code used in this paper publicly available.} Our dataset and code are released for research purposes only.

 \section{FAQs}
\begin{figure}[h]
    \centering   \includegraphics[width=\columnwidth]{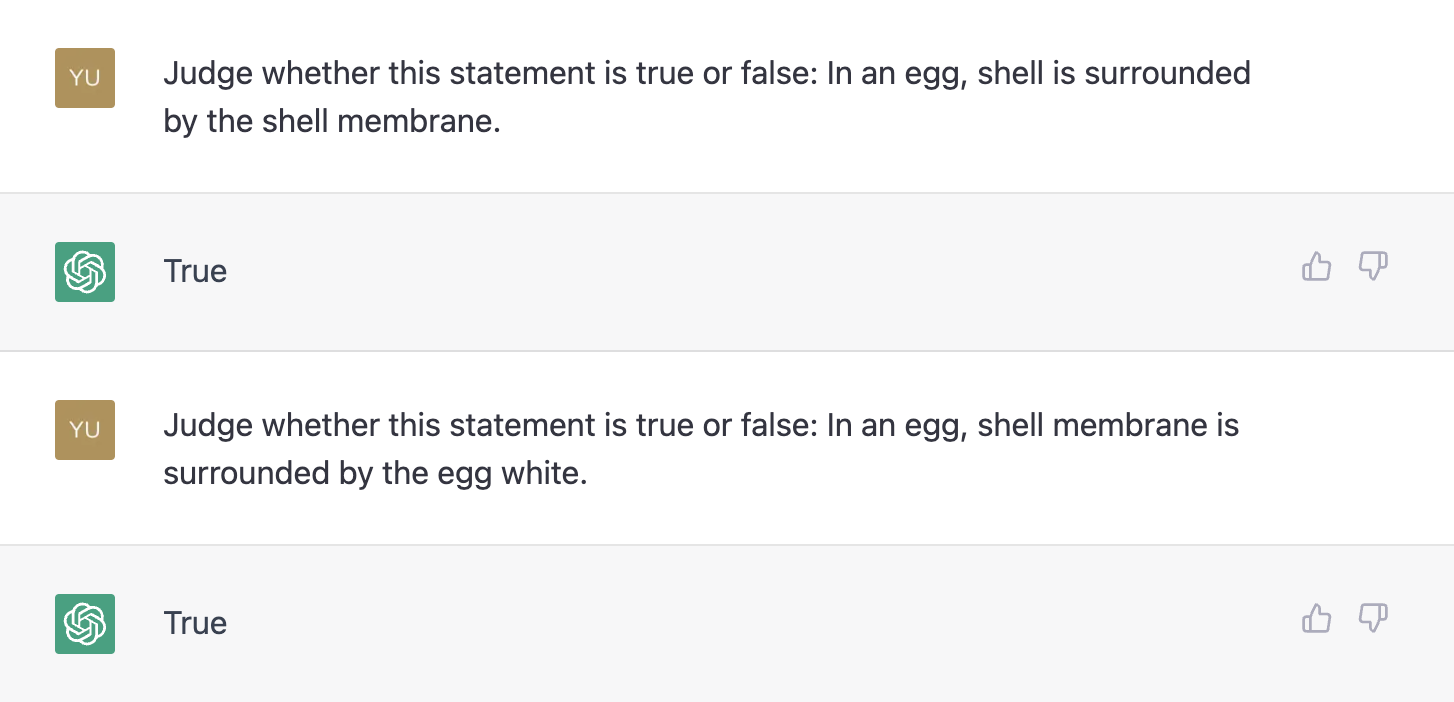}
         ChatGPT
    \caption{Like GPT-3 (text-davinci-003), ChatGPT also seems to have incoherent mental pictures of everyday things.
    \label{fig:chatgpt}}
\end{figure}

\begin{figure}[]
    \centering   \includegraphics[width=\columnwidth]{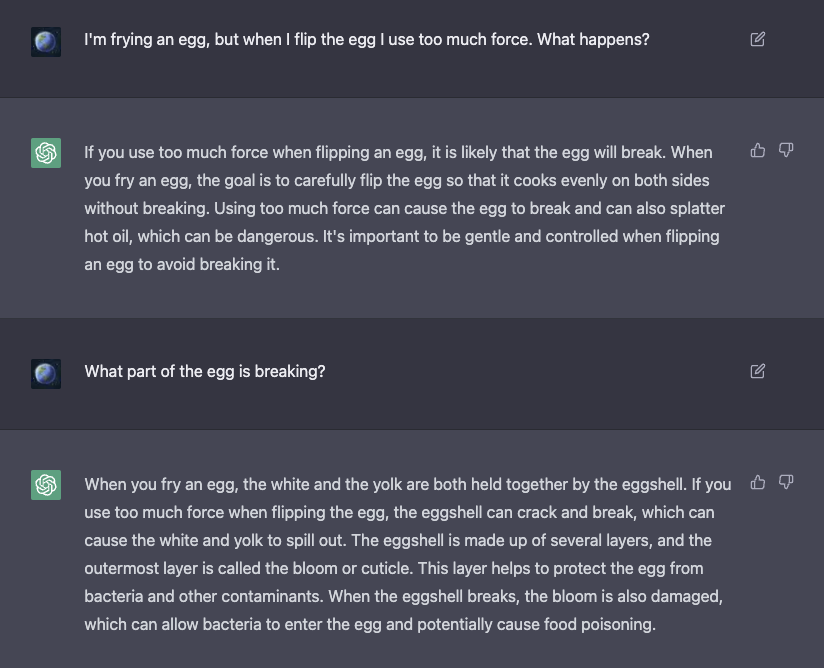}
         ChatGPT
    \caption{ChatGPT provides ridiculous responses regarding daily life activities such as frying an egg, illustrating poor mental models of everyday things and interactions with them. (Example by @bio\_bootloader, posted on Twitter \url{https://twitter.com/bio_bootloader/status/1599131249553330176/photo/1} at 11:59 AM Dec 3, 2022.)
    \label{fig:chatgpt_twitter}}
\end{figure}

 \begin{itemize}
     \item [\textbf{Q:}] \textbf{Does ChatGPT do better?}
     \item [] From informal tests, we find that ChatGPT is not devoid of mistakes either. We provide some examples to illustrate how the lack of coherent mental models of everyday things may also appear for other models of the GPT-3.5 family, like ChatGPT in Figure \ref{fig:chatgpt}. Others have also found ChatGPT responses that convey ridiculous interactions with everyday things e.g. it generates that ``When you fry an egg, the white and the yolk are both held together by the eggshell.'' (See Figure \ref{fig:chatgpt_twitter})

    \item [\textbf{Q:}] \textbf{GPT-3 and ChatGPT models are often updated, when were the models accessed for your experiments?}
    \item [] In our experiments with GPT-3, we used the text-davinci-003 model and queried the API on December 16, 2022 (during the period of time between 12 PM to 3.30 PM PST). 
    ChatGPT as in Figure \ref{fig:chatgpt} was accessed on December 17, 2022 (at around 9.30 PM PST). It would be interesting for researchers to investigate if future versions of the systems can construct better \mms{} of everyday things. \\
    \item [\textbf{Q:}] \textbf{How do you ensure high-quality mental models are acquired via crowdsourcing?}
    \item []
     We enforced a set of manual and automated checks during data acquisition which includes collecting mental model sketches and transcribing them into relation tuples. \\
     \textbf{Manual checks:} We randomly sampled 15 mental model sketches and made sure that the transcription of relation tuples was accurate i.e. all the relations tuples in mental model sketches drawn by crowdworkers were precisely added to our dataset. We also checked the quality and format of sketches (`.png' files) which will be released with our dataset.\\
    \textbf{Automated checks:} After enriching with implied relations, we also programatically checked that all individual mental models (total of 11.7K relations) in \dataset{} are fully consistent (based on the 4 commonsense constraints described in Section \ref{sec:constrained_reasoning}).

    \edit{
    \item [\textbf{Q:}] \textbf{Do similar trends apply to smaller models?}
    \item []
    Experiments on Macaw-3B, Macaw-large, UnifiedQA-large pointed towards the same trends. We also make our code and data fully accessible at \url{https://github.com/allenai/everyday-things} for interested researchers to experiment with other models of interest to them.
    }

    \edit{
    \item [\textbf{Q:}] \textbf{Can \method{} be applied to other languages?}
    \item []
    While our dataset is in English, relationships between parts of everyday things could indeed be authored for/ translated into other languages. We made our code and data publicly available, so others could use the infrastructure to apply the technique to other languages.
    }

 \end{itemize}

\end{document}